%% file: main.tex
\documentclass[10pt,twocolumn,letterpaper]{article}

\usepackage{cvpr}              

\usepackage{booktabs}
\usepackage{array}
\usepackage{multicol}
\usepackage{multirow}
\usepackage{amsmath}
\usepackage{breqn}
\usepackage{algorithm}
\usepackage{algpseudocode}

\usepackage{amsmath, amssymb}

 \usepackage{mathtools}
 \usepackage{colortbl}
 \usepackage{xcolor}

\usepackage[english]{babel}
\usepackage{amsthm}

\newtheorem{theorem}{Theorem}[section]

\newtheorem{lemma}[theorem]{Lemma}
\newtheorem{assumption}{Assumption}

\usepackage{pifont}

\definecolor{cvprblue}{rgb}{0.21,0.49,0.74}
\usepackage[pagebackref,breaklinks,colorlinks,allcolors=cvprblue]{hyperref}

\def\paperID{*****} 
\def\confName{CVPR}
\def\confYear{2026}

\title{The Geometry of Robustness: Optimizing Loss Landscape Curvature and Feature Manifold Alignment for Robust Finetuning of Vision-Language Models}

\author{Shivang Chopra \quad Shaunak Halbe \quad Chengyue Huang \quad Brisa Maneechotesuwan \quad Zsolt Kira\\
Georgia Institute of Technology\\
{\tt\small \{shivangchopra11,shalbe9,chuang475,bmaneech3,zkira\}@gatech.edu}
}

\begin{document}
\maketitle
\input{sec/0_abstract}    
\input{sec/1_intro}

\input{sec/2_relwork}
\input{sec/3_theoretical_analysis}

\input{sec/4_geometry}
\input{sec/5_method}
\input{sec/6_experiments}
\input{sec/7_conclusion}

{
    \small
    \bibliographystyle{ieeenat_fullname}
    \bibliography{main}
}

\input{sec/X_suppl}

\end{document}

%% file: sec/0_abstract.tex
\begin{abstract}
Fine-tuning approaches for Vision-Language Models (VLMs) face a critical three-way trade-off between In-Distribution (ID) accuracy, Out-of-Distribution (OOD) generalization, and adversarial robustness. Existing robust fine-tuning strategies resolve at most two axes of this trade-off. Generalization-preserving methods retain ID/OOD performance but leave models vulnerable to adversarial attacks, while adversarial training improves robustness to targeted attacks but degrades ID/OOD accuracy. Our key insight is that the robustness trade-off stems from two geometric failures: sharp, anisotropic minima in parameter space and unstable feature representations that deform under perturbation. To address this, we propose \textbf{GRACE (Gram-aligned Robustness via Adaptive Curvature Estimation)}, a unified fine-tuning framework that jointly regularizes the parameter-space curvature and feature-space invariance for VLMs. Grounded in Robust PAC-Bayes theory, GRACE employs adaptive weight perturbations scaled by local curvature to promote flatter minima, combined with a feature alignment loss that maintains representation consistency across clean, adversarial, and OOD inputs. On ImageNet fine-tuning of CLIP models, GRACE simultaneously improves ID accuracy by 10.8\%, and adversarial accuracy by 13.5\% while maintaining 57.0\% OOD accuracy (vs. 57.4\% zero-shot baseline). Geometric analysis confirms that GRACE converges to flatter minima without feature distortion across distribution shifts, providing a principled step toward generalized robustness in foundation VLMs.
\end{abstract}

%% file: sec/1_intro.tex
\vspace{-20pt}
\section{Introduction}
\label{sec:intro}
\vspace{-5pt}

Large-scale Vision-Language Models (VLMs) such as CLIP and ALIGN \cite{clip,align} have become powerful universal feature extractors, demonstrating strong zero-shot transfer and robustness to several natural distribution shifts \cite{basic}. However, when adapted to downstream tasks, their reliability is constrained by a persistent three-way trade-off between: (i) improving in-distribution (ID) accuracy, (ii) preserving robustness under natural and synthetic distribution shifts (OOD), and (iii) resisting gradient-based adversarial attacks. Standard fine-tuning often collapses at least one of these axes \cite{tpgm,ftp,fare,NEURIPS2023_a97b58c4}, making \emph{generalized robustness} (simultaneous robustness across ID, OOD, and adversarial regimes) a central open challenge for VLM deployment.

\begin{figure}
    \centering
    \includegraphics[width=\linewidth]{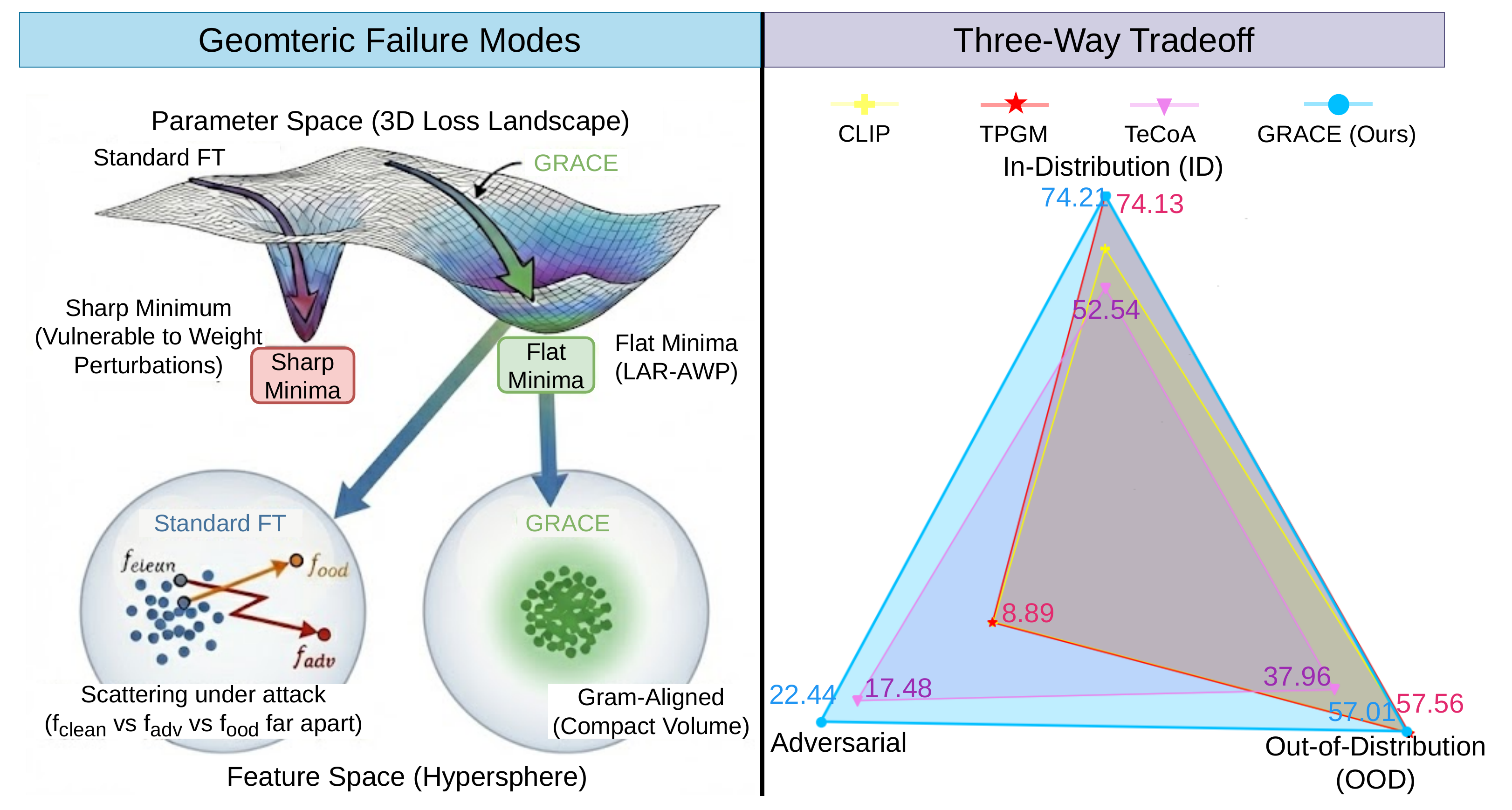}
    \caption{\textbf{The VLM robustness three-way tradeoff.} Existing robust fine-tuning strategies resolve at most two of \{ID, OOD, adversarial\} robustness simultaneously, leaving a gap in generalized robustness. GRACE is designed to close this gap. \vspace{-18pt}}
    \label{fig:spider}
\end{figure}

Empirically, we find that existing robust fine-tuning methods systematically \emph{polarize} this trade-off rather than resolve it. Two dominant strategies emerge:
\begin{enumerate}[label=\textbf{S\arabic*}, leftmargin=*]
    \item \textbf{Generalization-focused fine-tuning.} Methods that preserve or enhance zero-shot and OOD performance via conservative adaptation, weight regularization \cite{wiseft,tpgm,ftp,spd,digrap} or grounding with text anchors \cite{flyp, lipsumft} typically maintain strong ID/OOD accuracy, but exhibit near-zero robustness to standard $\ell_p$ PGD attacks.
    \item \textbf{Adversarial training for VLMs.} Conventional Adversarial Training (AT) approaches substantially improve PGD robustness but suffer notable drops in both ID and OOD accuracy. Critically, our analysis shows that current AT approaches \cite{tecoa, fare} result in a significant reduction in robustness to natural adversarial samples in datasets like ImageNet-A/A-Plus \cite{madry2018towards}.
\end{enumerate}

These behaviors indicate that the robustness trade-off is not merely a hyperparameter artifact, but reflects how different objectives reshape the underlying geometry of the model. In Sections~\ref{sec:theory},\ref{sec:geometry}, we perform a systematic theoretical and geometric analysis of CLIP-style VLMs under strategies S1 and S2, examining both the local loss landscape and the induced feature manifold. This analysis reveals three coupled failures: (\emph{i}) large deviation from pre-trained weights, (\emph{ii}) sharp, anisotropic minima associated with high parameter-space complexity, and (\emph{iii}) feature manifolds that deform under distribution shifts, destabilizing alignment across clean, OOD, and adversarial inputs.

Motivated by this perspective, we propose \textbf{GRACE (Gram-aligned Robustness via Adaptive Curvature Estimation)}, a unified fine-tuning framework grounded in a robust PAC-Bayesian perspective. GRACE is designed to jointly control:
(i) \emph{parameter-space curvature}, promoting flatter, lower-complexity solutions around the pre-trained weights; and
(ii) \emph{feature-space invariance}, encouraging stable class structure under both input and weight perturbations.
It instantiates these principles via two components:
\begin{enumerate}[leftmargin=*]
    \item \textbf{Layerwise Adaptive Low-Rank Adversarial Weight Perturbation (LAR-AWP):} structured, low-rank adversarial perturbations with layerwise adaptive magnitudes that bias optimization toward flatter regions without the severe accuracy penalties of standard AT;
    \item \textbf{Gram-Volume Alignment Loss:} a compact statistic capturing how image features spread in representation space, enforced to remain consistent across clean, LAR-AWP-perturbed, and adversarial-perturbed states, thereby preserving class separation and alignment with clean embeddings across domains and attacks.
\end{enumerate}

\noindent
This paper makes the following key contributions:
\begin{itemize}[leftmargin=*]
    \item We provide a theoretical and geometric analysis of the VLM fine-tuning robustness trade-off (Section~\ref{sec:theory}, \ref{sec:geometry}), linking failures of existing methods to sharp, anisotropic minima and feature manifold distortion across clean, OOD, and adversarial inputs.
    \item We propose \textbf{GRACE}, a principled geometry-driven framework that jointly regularizes parameter-space curvature and feature-space invariance under a robust PAC-Bayesian view for achieving generalized robustness.
    \item We demonstrate that GRACE achieves improved ID accuracy, OOD robustness, and adversarial robustness across multiple CLIP backbones and challenging benchmarks, yielding a more favorable robustness–accuracy profile than prior robust fine-tuning and AT approaches.
\end{itemize}

%% file: sec/2_relwork.tex
\section{Related Work}
\vspace{-5pt}
\textbf{Robust Fine-tuning of Vision-Language Models.}
Robust fine-tuning strategies for VLMs aim to retain pre-trained knowledge while adapting to new distributions. There are primarily two ways in which methods achieve this: (i) constrained adaptation by forcing the fine-tuned weights to remain close to the pre-trained model \cite{wiseft, tpgm, ftp, spd}, and (ii) grounding the image features into text anchors \cite{flyp, lipsumft}. However, these methods primarily target the ID–OOD axis of the trade-off—yielding flatter minima and better domain generalization—but remain vulnerable to gradient-based adversarial attacks and natural adversarial shifts.  Our proposed GRACE framework addresses this limitation by coupling constrained adaptation (LoRA) with curvature regularization and feature-space invariance. 

\textbf{Adversarial Fine-tuning of VLMs.} Adversarial training (AT)~\cite{madry2018towards} has proven effective for improving robustness to small input perturbations but often exacerbates OOD degradation. Existing approaches typically (i) preserve proximity to the pre-trained model~\cite{pmg_aft}, (ii) enforce invariance to pre-trained image features~\cite{fare}, or (iii) anchor image features to text~\cite{tecoa,laat}. While effective along the ID–Adv axis, these methods under-address the global geometry of the weight–loss landscape, which is critical for OOD stability and representational coherence. GRACE builds on invariance-based ideas~\cite{fare} by introducing a Gram-Volume regularizer that aligns clean, adversarial, and AWP-perturbed embeddings, and pairs it with curvature-aware adaptation to target the full ID–OOD–Adv trilemma.

\textbf{Generalization through Curvature and Geometry.}
There has been recent interest in improving generalization through adversarial training \cite{atda}. Recent research has shown that OOD samples could be interpreted as strong structured adversarial perturbations, utilizing low-rank universal structures in adversarial perturbations to improve model generalization under adversarial training \cite{structured}. Concurrently, research has also shown that regularizing the weight-loss landscape (i.e., sensitivity of loss to weight perturbations) can be very effective in promoting generalization \cite{sam, awp}. However, in large over-parameterized models, sharpness is highly anisotropic across layers and parameters, so uniform regularization is suboptimal. GRACE addresses this line by introducing \emph{Layerwise Adaptive Low-Rank Adversarial Weight Perturbation (LAR-AWP)}, which adaptively adjusts perturbation rank per layer based on local curvature, yielding targeted flattening where it matters.

%% file: sec/3_theoretical_analysis.tex
\section{Theoretical Analysis}
\label{sec:theory}
\vspace{1pt}
To understand the robustness trade-offs observed during fine-tuning, we start by theoretically deriving the necessary conditions for generalized robust fine-tuning of Vision-Language Models. Subsequently, we analyze the failure modes in existing methods and motivate GRACE to effectively balance the trade-off.

\subsection{Problem Setup and Assumptions}

Consider fine-tuning a pre-trained VLM with parameters $\theta_0 \in \mathbb{R}^d$ to parameters $\theta$ using dataset $\mathcal{D} = \{(x_i, y_i)\}_{i=1}^n$. Let $f_\theta: \mathcal{X} \rightarrow \mathbb{R}^k$ denote the model with loss $\ell: \mathbb{R}^k \times \mathcal{Y} \rightarrow \mathbb{R}$. For domain $s \in \mathcal{S} = \{\text{ID}, \text{OOD}, \text{Adv}\}$, define risk:
\begin{equation}
R_s(\theta) = \mathbb{E}_{(x,y) \sim \mathcal{D}_s}[\ell(f_\theta(x), y)]
\end{equation}
The robust risk is $R_{\text{Rob}}(\theta) = \max_{s \in \mathcal{S}} R_s(\theta)$.

\begin{assumption}[Smoothness]
\label{asmp:smooth}
The loss $\ell(f_\theta(\cdot), \cdot)$ is $L$-smooth and $\beta$-Lipschitz in $\theta$ with bounded Hessian spectrum: $\|\nabla^2_\theta \ell\| \leq M$.
\end{assumption}

\begin{assumption}[Feature Regularity]
\label{asmp:features}
The encoder $f_\theta$ maps to the unit sphere with $L_f$-Lipschitz continuity: $\|f_\theta(x) - f_{\theta'}(x)\| \leq L_f\|\theta - \theta'\|$ for all $x$.
\end{assumption}

\subsection{PAC-Bayesian Framework for Multi-Domain Robustness}

Following the PAC-Bayes framework \cite{mcallester2013pacbayesiantutorialdropoutbound, neyshabur2018pacbayesianapproachspectrallynormalizedmargin}, we model fine-tuning as sampling from posterior $Q = \mathcal{N}(\theta, \sigma^2 I)$ given prior $P = \mathcal{N}(\theta_0, \sigma^2 I)$.

\begin{theorem}[Robust PAC-Bayes Bound]
\label{thm:main}
Under Assumptions \ref{asmp:smooth}-\ref{asmp:features}, with probability at least $1-\delta$ over the training set:
\begin{align}
R_{\text{Rob}}(\theta) \leq \hat{R}_{\text{ID}}(\theta) &+ \underbrace{\frac{\|\theta - \theta_0\|^2}{2n\sigma^2} + \frac{\ln(2n/\delta)}{2n}}_{\text{(A) Proximity from prior}} \nonumber \\
&+ \underbrace{\frac{\sigma^2}{2} \cdot \text{Tr}(\mathbb{E}[\nabla^2_\theta R_{\text{Rob}}(\theta)])}_{\text{(B) Parameter-space sharpness}} \nonumber  \\
&+ \underbrace{\max_{s,t \in \mathcal{S}} d_{\mathcal{H}\Delta\mathcal{H}}(\mathcal{D}_s, \mathcal{D}_t)}_{\text{(C) Cross-domain discrepancy}} + \lambda^*  \label{eqn:main}
\end{align}
where $\hat{R}_{\text{ID}}$ is empirical ID risk, $d_{\mathcal{H}\Delta\mathcal{H}}$ is the $\mathcal{H}\Delta\mathcal{H}$-divergence, and $\lambda^*$ is the ideal joint error.
\end{theorem}
Refer to the supplementary for the proof of Theorem \ref{thm:main}.

\subsection{Geometric Interpretation and Feature-Space Bounds}

The $\mathcal{H}\Delta\mathcal{H}$-divergence in term (C) is intractable for neural networks. We derive a feature-space upper bound:

\begin{lemma}[Feature-Space Domain Discrepancy]
\label{lem:feature}
Under Assumption \ref{asmp:features}, the domain divergence is bounded by class-conditional feature alignment:
\begin{equation}
d_{\mathcal{H}\Delta\mathcal{H}}(\mathcal{D}_s, \mathcal{D}_t) \leq 2L_f \sum_{c=1}^k \pi_c \left( \|\mu_s^c - \mu_t^c\|_2 + \sqrt{\text{Tr}(\Sigma_s^c - \Sigma_t^c)^2} \right)
\end{equation}
where $\mu_s^c, \Sigma_s^c$ are the mean and covariance of class-$c$ features under domain $s$, and $\pi_c$ is the class prior.
\end{lemma}

This connects domain discrepancy to observable geometric properties: class centroid alignment and covariance stability.

\begin{table}[h]
\centering
\resizebox{\linewidth}{!}{
\begin{tabular}{lccc}
\toprule
\textbf{Method} & \textbf{(A) Proximity} & \textbf{(B) Sharpness} & \textbf{(C) Stability} \\
& $\|\theta - \theta_0\|^2$ & $\text{Tr}(\nabla^2 R)$ & $d_{H\Delta H}$ \\
\midrule
Vanilla FT & {\color{red}\ding{55}} & {\color{red}\ding{55}} & {\color{red}\ding{55}} \\
WiSE-FT \cite{wiseft} & \checkmark & {\color{red}\ding{55}} & {\color{red}\ding{55}} \\
FLYP \cite{flyp} & \checkmark & {\color{red}\ding{55}} & $\sim$ \\
TPGM \cite{tpgm} & \checkmark & {\color{red}\ding{55}} & {\color{red}\ding{55}} \\
SPD \cite{spd} & \checkmark & {\color{red}\ding{55}} & {\color{red}\ding{55}} \\
\midrule
TeCoA \cite{tecoa} & {\color{red}\ding{55}} & $\sim$ & $\sim$ \\
FARE \cite{fare} & {\color{red}\ding{55}} & $\sim$ & \checkmark \\
PMG-AFT \cite{pmg_aft} & \checkmark & $\sim$ & \checkmark \\
LAAT \cite{laat} & {\color{red}\ding{55}} & \checkmark & \checkmark \\
\midrule
\rowcolor{gray!15}
\textbf{GRACE (Ours)} & \checkmark & \checkmark & \checkmark \\
\bottomrule
\end{tabular}
}
\vspace{-0.5em}
\caption{Decomposition of robust fine-tuning methods by optimization objectives. Methods addressing each term explicitly through their loss function or training procedure are marked (\checkmark), those with implicit or partial coverage are marked ($\sim$), and those without explicit optimization are marked ({\color{red}\ding{55}}). \vspace{-10pt}}
\label{tab:method_objectives}
\vspace{-1em}
\end{table}

\begin{figure*}
    \centering
    \vspace{-10pt}
    \includegraphics[width=0.9\linewidth]{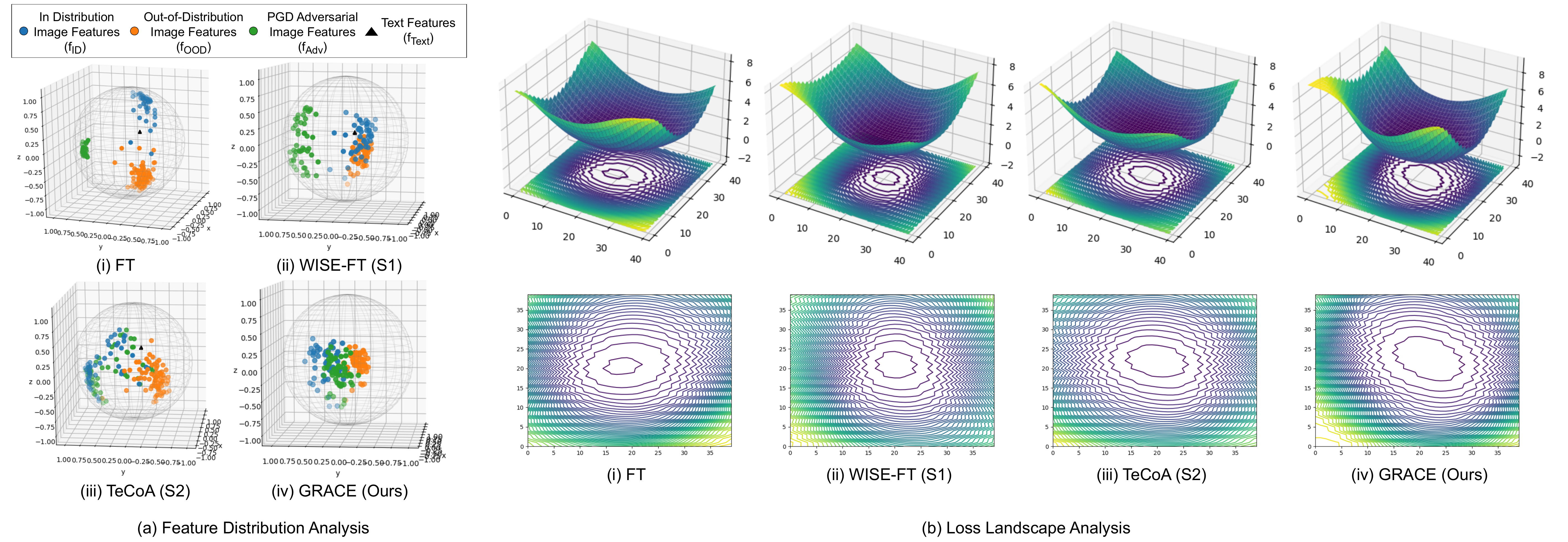}
    \caption{(a) \textbf{Feature Distribution Analysis}: 3D projection of image features for in-distribution ($f_{\text{ID}}$), OOD ($f_{\text{OOD}}$), and PGD adversarial inputs ($f_{\text{Adv}}$) of the same class. (b) \textbf{Loss Landscape Analysis}: 3D/2D loss slices around the converged solutions for each method, using shared perturbation directions. \vspace{-20pt}}
    \label{fig:intro_analysis}
\end{figure*}

\subsection{Predicted Failure Modes from Incomplete Optimization}

Theorem \ref{thm:main} reveals that robust fine-tuning requires simultaneous control of three critical terms: (A) proximity to the pre-trained prior $\|\theta - \theta_0\|^2$, (B) parameter-space sharpness $\text{Tr}(\nabla^2 R_{\text{Rob}})$, and (C) feature-space stability $d_{H\Delta H}(\mathcal{D}_s, \mathcal{D}_t)$. We hypothesize that methods optimizing only a subset of these objectives will exhibit predictable failure modes: missing (A) leads to drift from the pre-trained manifold and degraded zero-shot transfer; missing (B) results in sharp minima vulnerable to adversarial perturbations; and missing (C) causes unstable feature representations that degrade under distribution shifts. Table \ref{tab:method_objectives} categorizes existing methods by which objectives they explicitly address. This decomposition predicts specific failure patterns validated in our experiments. Proximity-preserving methods like WiSE-FT/TPGM maintain zero-shot capabilities but remain adversarially vulnerable due to sharp, anisotropic landscape. Adversarial methods like FARE improve robustness by ensuring feature stability but drift away from $\theta_0$, harming OOD accuracy, while methods like TeCoA harm OOD the most since they don't explicitly optimize any of these objectives.

%% file: sec/4_geometry.tex
\begin{table}[t]
    \centering
    \small

    \label{tab:geometry_metrics}
        \centering
        \resizebox{\linewidth}{!}{
        \begin{tabular}{lcccc}
            \toprule
            & \multicolumn{2}{c}{Cosine Alignment} & \multicolumn{2}{c}{$\Delta$LID (↓)} \\
            \cmidrule(lr){2-3} \cmidrule(lr){4-5}
            Method & ID$\to$OOD & ID$\to$Adv & ID$\to$OOD & ID$\to$Adv \\
            \midrule
            FT          & 0.84 & 0.48 & +2.7 & +6.5 \\
            WiSE-FT (S1)& \textbf{0.91} & 0.57 & \textbf{+1.1} & +5.6 \\
            TeCoA (S2)  & 0.78 & 0.72 & +4.7 & +2.9 \\
            \textbf{GRACE} & 0.89 & \textbf{0.85} & +1.8 & \textbf{+2.5} \\
            \bottomrule
        \end{tabular}
        }
        \caption{\textbf{Class-conditional feature-space stability.}
        Cosine similarity between ID and shifted class centroids (left) 
        and change in LID relative to ID (right). Lower $\Delta$LID implies more stable local manifolds. \vspace{-15pt}}
        \label{tab:lid}
\end{table}

\section{Empirical Validation of Failure Modes}
\label{sec:geometry}

In this section, We validate the theoretical predictions from Section \ref{sec:theory} by demonstrating that each missing objective leads to its predicted failure mode. We fine-tune CLIP ViT-B/32 on ImageNet using methods from Table~\ref{tab:method_objectives}, evaluating on: (i) in-distribution (ID), (ii) out-of-distribution (OOD), and (iii) PGD-$\ell_\infty$ adversarial ($\epsilon{=}4/255$). We extract penultimate-layer embeddings normalized to the unit hypersphere, with metrics averaged over 100 samples per class across 3 seeds.

\vspace{-15pt}
\paragraph{Failure Mode 1: Loss of Proximity $\rightarrow$ Zero-Shot Degradation}
Methods without proximity control drift significantly from pre-trained weights. Measuring relative parameter distance $\|\theta - \theta_0\|_F / \|\theta_0\|_F$:
\textbf{High drift}: TeCoA (0.47), FARE (0.43), LAAT (0.41), FT (0.38);
\textbf{Low drift:} WiSE-FT (0.08), TPGM (0.11), PMG-AFT (0.13), GRACE (0.09).
This drift correlates strongly with zero-shot degradation (Pearson $r = -0.82$): methods exceeding 0.30 relative distance show $>$15\% drop in zero-shot accuracy (TeCoA: 38.75\% vs WiSE-FT: 60.40\%), confirming proximity preservation is critical for pre-trained knowledge retention.

\begin{table}[t]
    \centering
    \begin{tabular}{lcc}
        \toprule
        Method & $\lambda_{\max}\,(\times 10^{4})$ & $\|H\|_F/\sqrt{d}~(\times 10^{2})$ \\
        \midrule
        FT          & 3.5 & 0.89 \\
        WiSE-FT (S1)& 3.3 & 0.78 \\
        TeCoA (S2)  & 1.8 & 0.52 \\
        \textbf{GRACE} & \textbf{1.6} & \textbf{0.43} \\
        \bottomrule
    \end{tabular}
    \caption{\textbf{Parameter-space sharpness.} 
    Top Hessian eigenvalue and normalized Frobenius norm. \vspace{-15pt}}
    \label{tab:hessian}
\end{table}

\vspace{-15pt}
\paragraph{Failure Mode 2: Sharp Minima $\rightarrow$ Adversarial Vulnerability} Table~\ref{tab:hessian} shows methods lacking sharpness control converge to high-curvature regions. WISE-FT exhibits $\lambda_{\max} = 3.3 \times 10^3$, nearly double that of TeCoA (1.8 $\times 10^3$). Figure~\ref{fig:intro_analysis}(b) visualizes these differences through loss landscapes---unregularized methods show sharp, anisotropic minima while GRACE maintains moderate curvature (1.6 $\times 10^3$) comparable to WiSE-FT. Critically, sharpness correlates with adversarial vulnerability: WiSE-FT achieves 0\% adversarial accuracy, while GRACE's joint optimization yields 22.4\% robustness.

\paragraph{Failure Mode 3: Feature Instability $\rightarrow$ OOD Degradation}  
\vspace{-10pt}
Figure~\ref{fig:intro_analysis}(a) reveals how feature representations collapse under distribution shifts. WiSE-FT preserves ID/OOD structure (0.91 cosine alignment) but PGD features scatter from class manifolds (0.57 alignment), with $\Delta$LID increasing by +5.6. TeCoA maintains adversarial alignment (0.72) but fragments OOD manifolds ($\Delta$LID +4.7). Table~\ref{tab:lid} quantifies this: only GRACE maintains both high alignment (ID$\to$OOD: 0.89, ID$\to$Adv: 0.85) and low dimensionality increase ($\Delta$LID $\leq$ 2.5), indicating stable representations across all shifts.
\vspace{-10pt}
\paragraph{Summary: Joint Optimization is Necessary}
Our analysis confirms the theoretical predictions: proximity loss causes zero-shot degradation (TeCoA: 38.75\% vs WiSE-FT: 60.40\%), uncontrolled sharpness leads to adversarial vulnerability (WiSE-FT: 0\% robust accuracy), and feature instability degrades OOD performance (TeCoA: 37.96\% OOD avg).

%% file: sec/5_method.tex
\section{Methodology: The GRACE Framework}
\label{sec:method}

\subsection{Overview}

Building on the PAC-Bayesian view in Section~\ref{sec:theory},
we introduce GRACE
(\textbf{G}ram-aligned \textbf{R}obustness via \textbf{A}daptive \textbf{C}urvature \textbf{E}stimation),
a unified fine-tuning framework that jointly regularizes
\emph{parameter-space sharpness} and \emph{feature-space stability}
for CLIP-style VLMs.

Recalling the composite bound in Theorem~\ref{thm:main},
robust risk is controlled by:
(i) empirical ID fit,
(ii) expected robust sharpness,
(iii) proximity to the pre-trained prior (KL),
and (iv) pairwise domain discrepancy across ID/OOD/Adv.
GRACE instantiates these terms via three modules:

\begin{itemize}[leftmargin=10pt, itemsep=2pt, topsep=2pt]
    \item \textbf{LoRA fine-tuning} (prior proximity): parameter-efficient adaptation
          in a low-rank subspace anchored at the pre-trained model.
    \item \textbf{Layer-wise Adaptive Low-Rank AWP (LAR-AWP)} (sharpness control):
          adversarial weight perturbations in the LoRA subspace, with rank adapted per layer based on local curvature.
    \item \textbf{Gram-volume feature alignment} (domain discrepancy control):
          a Gram-volume-based loss that stabilizes per-sample representations
          across clean, adversarial, and weight-perturbed regimes.
\end{itemize}

The overall training objective is
\begin{equation}
\mathcal{L}_{\text{GRACE}}
=
\mathcal{L}_{\text{task}}
+
\lambda_{\text{LAR}}\,\mathcal{L}_{\text{LAR-AWP}}
+
\lambda_{\text{GV}}\,\mathcal{L}_{\text{GV}},
\label{eq:loss-grace}
\end{equation}
where $\mathcal{L}_{\text{task}}$ is the standard cross-entropy loss,
$\mathcal{L}_{\text{LAR-AWP}}$ controls robust sharpness,
and $\mathcal{L}_{\text{GV}}$ enforces feature invariance. We next describe each component of the proposed GRACE framework.

\begin{figure}[t]
    \centering
    \includegraphics[width=0.7\linewidth]{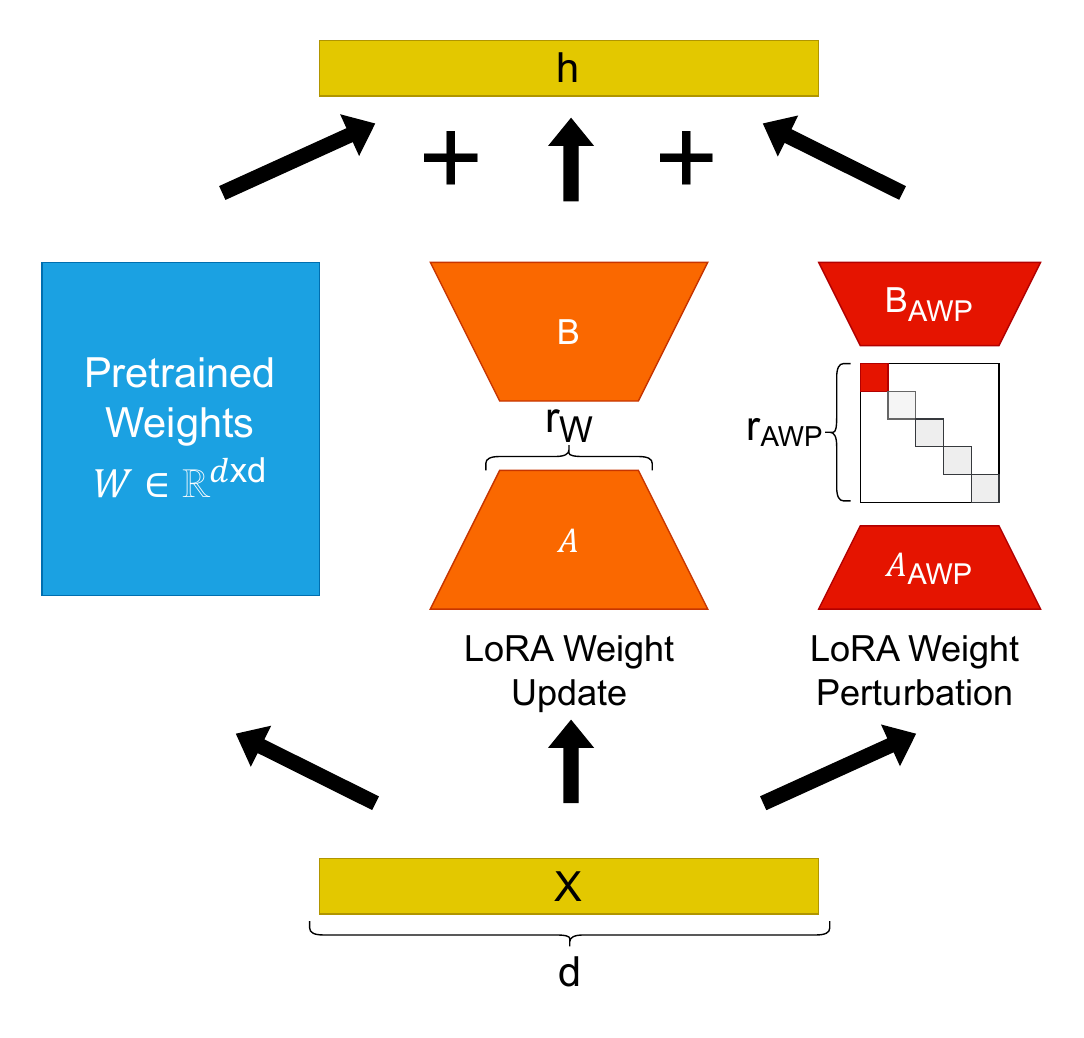}
    \caption{\textbf{Low-rank adaptation and perturbation in GRACE.}
    Frozen pretrained weights $W(\theta_0)$ (blue) are updated only through
    low-rank LoRA adapters (orange).
    LAR-AWP injects additional low-rank perturbations (red) in the same subspace.\vspace{-15pt}}
    \label{fig:lora_schematics}
\end{figure}

\subsection{LoRA Fine-Tuning: Prior Proximity}
\label{subsec:lora-method}

We adapt only low-rank adapters on top of the frozen CLIP backbone.
For each weight matrix $W \in \mathbb{R}^{n_1\times n_2}$,
\begin{equation}
W(\theta)
=
W(\theta_0) + B_W A_W,
\qquad
A_W \in \mathbb{R}^{r \times n_2},\;
B_W \in \mathbb{R}^{n_1 \times r},
\label{eq:lora-method}
\end{equation}
with rank $r \ll \min(n_1,n_2)$.
Only $\Theta=\{A_W,B_W\}_W$ is trainable; $W(\theta_0)$ is frozen.
This low-rank parameterization constrains the adapted parameters to remain in a small
affine subspace around the pre-trained weights, effectively controlling the KL term
$\mathrm{KL}(Q\|P)$ in Eq.~\eqref{eqn:main}.
We use the same LoRA configuration across the PEFT baselines and GRACE for fairness.

\subsection{Layer-Wise Adaptive Low-Rank AWP (LAR-AWP)}
\label{subsec:lar-awp}

To control the expected robust sharpness term in Eq.~\eqref{eqn:main},
GRACE injects adversarial weight perturbations in the same low-rank subspace, analogous to AWP~\cite{awp} but adapted to LoRA and made layer-wise rank-adaptive. The layerwise adaptive rank is motivated by the anisotropic structure of $\nabla^2 R$. The Hessian $\nabla^2 R$ exhibits layer-dependent eigenvalue gaps in VLMs. Uniform perturbation is suboptimal; allocating perturbation rank proportional to $\lambda_{\max}^{(\ell)}$ (layer $\ell$'s maximum eigenvalue) yields better complexity-smoothness tradeoffs.

\textbf{Low-rank perturbation parameterization.} For each LoRA-parameterized layer in Eq.~\eqref{eq:lora-method}, we introduce a perturbation branch
\begin{equation}
W_{\text{pert}}(\theta,\Delta)
=
W(\theta_0) + B_W A_W + B_{AWP} A_{AWP},
\label{eq:lora-awp-param}
\end{equation}
where $A_{AWP} \in \mathbb{R}^{r_{\text{awp},W} \times n_2}$ and
$B_W^{\text{pert}} \in \mathbb{R}^{n_1 \times r_{AWP}}$
are adversarial perturbation factors for layer $W$,
and $r_{AWP}$ is a layer-specific perturbation rank.
Figure~\ref{fig:lora_schematics} illustrates how LoRA updates (orange) and LAR-AWP perturbations (red) share the same low-rank geometry.

\begin{figure}[t]
    \centering
    \includegraphics[width=\linewidth]{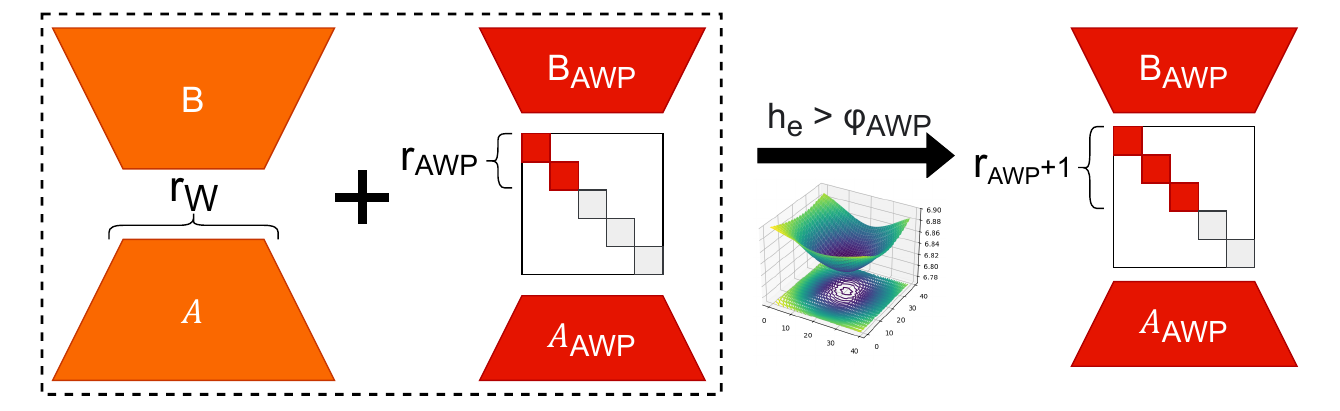}
    \caption{\textbf{LAR-AWP rank curriculum.}
    A diagonal rank mask controls the effective perturbation rank per layer.
    Curvature estimates $h_W$ (from mini-batch gradients) are used to assign higher
    perturbation ranks to sharper layers, focusing smoothing where the loss landscape
    is steepest. \vspace{-15pt}}
    \label{fig:curriculum}
\end{figure}

\textbf{Rank-adaptive curvature curriculum.}
Rather than using a fixed rank for all layers,
LAR-AWP assigns higher perturbation ranks to layers with larger curvature. The curvature of the loss function can be determined by its Hessian \cite{sophia}. Therefore, GRACE adapts the rank of AWPs for each parameter based on the Hessian trace for that parameter with respect to the loss function. However, computing the exact Hessian for large-scale models is computationally prohibitive. To mitigate this issue, Sophia \cite{sophia} proposed a method to estimate the trace of the diagonal of the Hessian matrix using mini-batch gradients. Specifically, the first-order gradients $g$ of the cross-entropy loss function $\mathcal{L}_{ce}$ are used to get an unbiased estimate for the diagonal of the Gauss-Newton matrix, which is a biased estimator for the trace of the Hessian, $h$.

\[
g_W = \nabla_W \frac{1}{n_v} \sum_{i=1}^{n_v} \mathcal{L}(F_{\theta}(x_{v_i}),y_{v_i}),
\qquad
h_W \approx n_v \, g_W \odot g_W,
\]
where $n_v$ is the validation mini-batch size.
We maintain an exponential moving average of $h_W$ for each layer and compute
a scalar curvature score per layer (e.g., mean of $h_W$).
Based on these scores, we allocate perturbation ranks $r_{\text{AWP}}$
according to curvature percentiles:
layers in the highest percentile receive the largest rank,
while flat layers are perturbed with minimal rank.
This curriculum is illustrated in Figure~\ref{fig:curriculum}.

\textbf{LAR-AWP objective.} At each training step, we perform an inner maximization over the weight perturbations, approximated by gradient-ascent steps in the low-rank perturbation space:
\begin{equation}
\mathcal{L}_{\text{LAR-AWP}}
\approx
\frac{1}{n}\sum_{i=1}^n
\max_{\substack{\|\delta_i\|_p \le \epsilon \\ \|\Delta\|\le \rho}}
\mathcal{L}\big(F_{W_{\text{pert}}(\theta,\Delta)}(x_i), y_i\big),
\label{eq:lar-awp-loss}
\end{equation}
where $\rho$ bounds the perturbation in the low-rank space.
The outer update then minimizes
$\mathcal{L}_{\text{task}} + \lambda_{\text{LAR}}\mathcal{L}_{\text{LAR-AWP}}$
with respect to $\Theta$.
This forces the model to perform well not only at $\theta$ but also in a local neighborhood defined by the weight perturbations, driving the solution toward flatter, more robust minima (consistent with the curvature trends in Table~\ref{tab:hessian}).

\subsection{Gram-Volume Alignment Loss}
\label{subsec:gram-loss}

The remaining term in Eq.~\eqref{eqn:main},
$\max_{s\neq t} d_{\mathcal{H}\Delta\mathcal{H}}(\mathcal{D}_s,\mathcal{D}_t)$,
captures pairwise domain discrepancy.
Exact $\mathcal{H}\Delta\mathcal{H}$-divergences are intractable for VLMs, so we operate in feature space and use Gram volume as a differentiable surrogate
for local feature spread.

\begin{figure}[t]
    \centering
    \includegraphics[width=\linewidth]{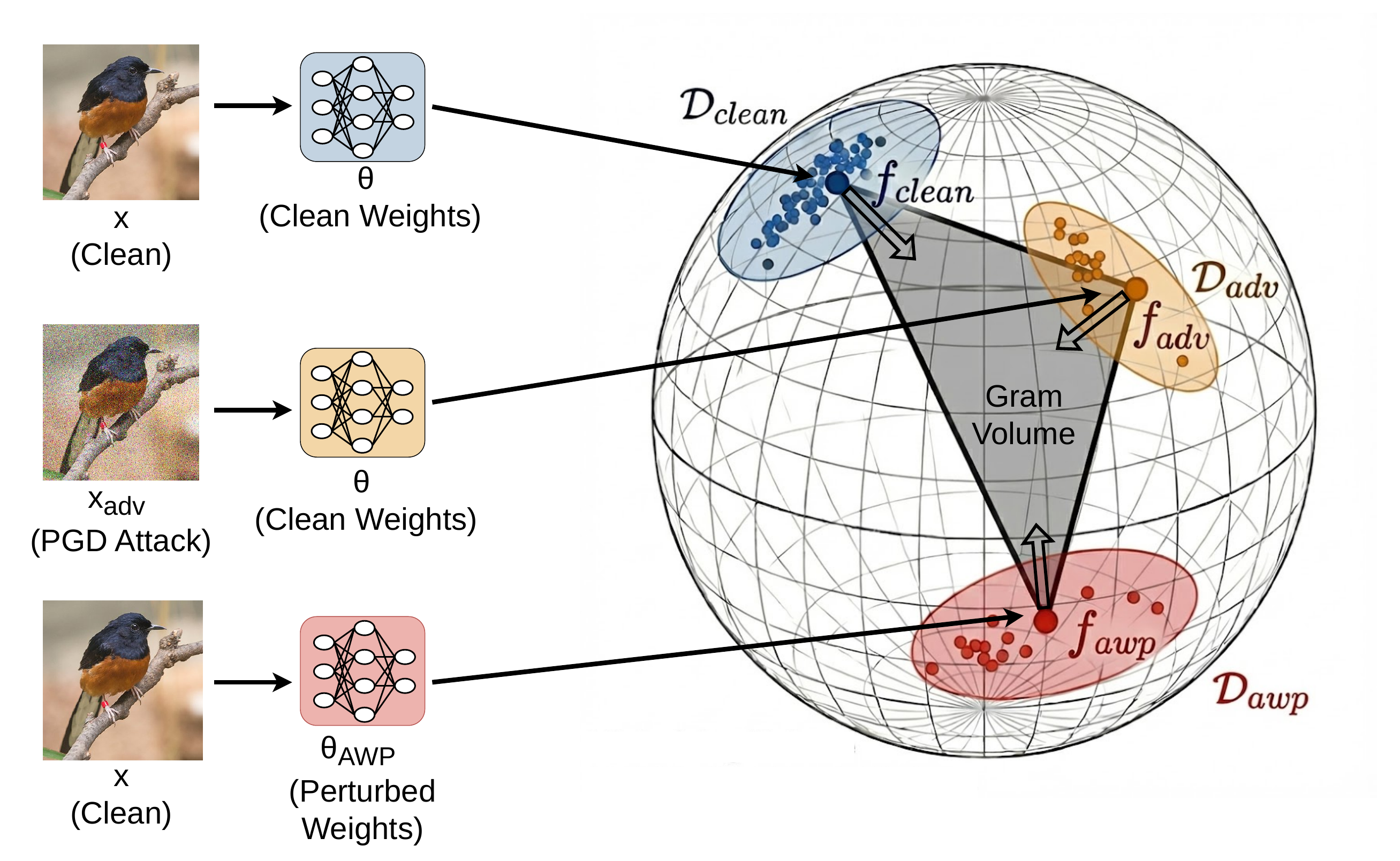}
    \caption{\textbf{Gram-volume feature alignment.}
    For each input, GRACE compares clean, adversarial, and LAR-AWP-perturbed
    image embeddings via a small Gram matrix.
    The Gram-volume loss encourages these three vectors to remain close to each other
    (low volume) while preserving separation across different classes. \vspace{-20pt}}
    \label{fig:gram_loss}
\end{figure}

Let $f_{\text{ID}}, f_{\text{Adv}}, f_{\text{AWP}} \in \mathbb{R}^D$ be
the $\ell_2$-normalized image features for sample $i$ under clean, adversarial, and
LAR-AWP-perturbed weights, respectively.
We define a $3\times 3$ Gram matrix
\begin{equation}
G_i =
\begin{bmatrix}
\langle f_{\text{ID}}, f_{\text{ID}} \rangle &
\langle f_{\text{ID}}, f_{\text{Adv}} \rangle &
\langle f_{\text{ID}}, f_{\text{AWP}} \rangle \\
\langle f_{\text{Adv}}, f_{\text{ID}} \rangle &
\langle f_{\text{Adv}}, f_{\text{Adv}} \rangle &
\langle f_{\text{Adv}}, f_{\text{AWP}} \rangle \\
\langle f_{\text{AWP}}, f_{\text{ID}} \rangle &
\langle f_{\text{AWP}}, f_{\text{Adv}} \rangle &
\langle f_{\text{AWP}}, f_{\text{AWP}} \rangle
\end{bmatrix}
+ \varepsilon I,
\label{eq:gram-matrix}
\end{equation}
where $\langle\cdot,\cdot\rangle$ is the inner product
and $\varepsilon$ is a small constant for numerical stability.
The associated Gram volume is
\begin{equation}
\mathcal{L}_{\text{GV}} = \sqrt{|\det(G_i)|}.
\label{eq:gram-vol}
\end{equation}
As illustrated in Figure~\ref{fig:gram_loss},
$\mathcal{L}_{\text{GV}}$ measures the volume of the parallelepiped
spanned by $\{f_{\text{ID}}, f_{\text{Adv}}, f_{\text{AWP}}\}$:
$\mathcal{L}_{\text{GV}} \approx 0$ when the three representations are close to each other (stable manifold),
and $\mathcal{L}_{\text{GV}}$ grows as perturbations push features into diverging directions.

\subsection{Training Procedure}

In practice, GRACE alternates the following steps per minibatch:
(i) compute clean features and $\mathcal{L}_{\text{task}}$;
(ii) generate PGD adversarial examples;
(iii) perform a few inner steps of low-rank weight perturbation (LAR-AWP),
guided by the curvature-based rank curriculum;
(iv) compute the Gram-alignment loss $\mathcal{L}_{\text{GV}}$
from clean/Adv/AWP features; and
(v) update the LoRA parameters $\Theta$ using the combined loss
in Eq.~\eqref{eq:loss-grace}.

%% file: sec/6_experiments.tex
\begin{table*}[t]
    \centering
    \tabcolsep=3pt
    \extrarowheight=2pt
    \resizebox{0.8\linewidth}{!}{
    \begin{tabular}{c | c  c | c  c | c  c | c  c | c  c | c  c | c  c | c  c }
    \toprule
         & \multicolumn{2} {c|} {\textbf{ID}} & \multicolumn{6} {c|} {\textbf{Domain Shift}} & \multicolumn{4} {c|} {\textbf{Natural Adversarial}} & \multicolumn{4} {c} {\textbf{Statistics}}\\
         \cmidrule(l{0.6em}r{0.6em}){4-9} \cmidrule(l{0.6em}r{0.6em}){10-13} \cmidrule(l{0.6em}r{0.6em}){14-17}
         \textbf{Method}  & \multicolumn{2} {c|} {\textbf{ImageNet}} & \multicolumn{2} {c|} {\textbf{ImageNet-V2}} & \multicolumn{2} {c|} {\textbf{ImageNet-S}} & \multicolumn{2} {c|} {\textbf{ImageNet-R}} & \multicolumn{2} {c|} {\textbf{ImageNet-A}} & \multicolumn{2} {c|} {\textbf{ImageNet-A-Plus}} & \multicolumn{2} {c|} {\textbf{OOD Avg.}}  & \multicolumn{2} {c} {\textbf{Nat Adv Avg}}\\
          
         & Clean & Adv & Clean & Adv & Clean & Adv & Clean & Adv & Clean & Adv & Clean & Adv & Clean & Adv & Clean & Adv \\
    \toprule
    \midrule
     CLIP & 63.35 & 0.00 & 56.48 & 0.00 & 41.44 & 0.00 & 68.83 & 0.01 & 32.69 & 0.00 & 37.92 & 0.00 & 55.58 & 0.00 & 35.30 & 0.00 \\
     \cmidrule{1-17}\noalign{\vskip 0.5ex}
     Vanilla FT & 74.86 & 0.00 & 63.75 & 0.00 & 43.34 & 0.00 & 60.85 & 0.00 & 22.47 & 0.00 & 29.12 & 0.00 & 55.98 & 0.00 & 25.80 & 0.00 \\
     WISE-FT & 70.20 & 0.00 & 61.06 & 0.00 & 43.34 & 0.00 & 63.78 & 0.00 & 32.60 & 0.00 & 39.74 & 0.00 & 56.06 & 0.00 & \textbf{36.17} & 0.00 \\

      FLYP & 72.15 & 0.00 & 62.22 & 0.00 & 42.22 & 0.00 & 64.38 & 0.01 & 30.06 & 0.00 & 37.89 & 0.00 & 56.27 & 0.00 & 33.98 & 0.00 \\

      TPGM & 74.13 & 0.00 & 63.20 & 0.00 & 42.71 & 0.00 & 64.79 & 0.00 & 32.38 & 0.00 & 38.73 & 0.00 & \textbf{56.90} & 0.00 & \underline{35.56} & 0.00 \\
      SPD & 73.68 & 0.00 & 62.60 & 0.00 & 42.56 & 0.00 & 64.60 & 0.00 & 32.07 & 0.00 & 37.46 & 0.00 & \underline{56.59} & 0.00 & 34.76 & 0.00 \\
      \cmidrule{1-17}\noalign{\vskip 0.5ex}

      TeCoA & 52.54 & 24.40 & 41.10 & 17.71 & 25.89 & 13.06 & 44.54 & 21.01 & 5.11 & 0.66 & 9.86 & 1.07 & 37.18 & 17.26 & 7.48 & 0.86 \\
      FARE & 47.38 & 13.94 & 39.00 & 9.77 & 28.63 & 11.05 & 47.79 & 17.05 & 5.97 & 0.54 & 10.38 & 0.68 & 38.47 & 12.62 & 8.18 & 0.61 \\
      PMG-AFT & 58.20 & 26.80 & 43.50 & 19.70 & 31.10 & 14.80 & 50.30 & 22.10 & 8.22 & 1.12 & 13.07 & 1.94 & 41.63 & \underline{18.87} & 10.64 & 1.53 \\
      LAAT & 55.46 & 21.54 & 42.30 & 15.80 & 30.15 & 12.20 & 48.90 & 20.40 & 12.23 & 1.93 & 16.48 & 3.48 & 40.45 & 16.13 & 14.35 & \underline{2.70} \\

      \midrule
      \rowcolor{gray!15}
      GRACE & 74.21 & 25.44 & 62.60 & 20.51 & 42.11 & 17.75 & 58.53 & 23.39 & 20.02 & 3.32 & 25.33 & 4.86 & 54.41 & \textbf{20.55} & 22.67 & \textbf{4.09} \\
   
    \bottomrule

    \end{tabular}
    }
    \vspace{-10pt}
    \caption{\textbf{OOD Results on ImageNet.} CLIP ViT-B/32 finetuned on ImageNet \cite{imagenet} dataset and evaluated on ImageNet variants. The numbers are top-1 accuracy (\%). OOD Avg averages ImageNet-V2, -S, -R; Nat Adv Avg averages ImageNet-A and A-Plus. \vspace{-10pt}}
    \label{tab:imagenet_vit_b_32}
\end{table*}

\begin{table*}[]
    \centering
    \resizebox{\linewidth}{!}{
    \begin{tabular}{c|cc|cc|cc|cc|cc|cc|cc|cc|cc}
    \toprule
         & \multicolumn{16} {c|} {\textbf{Zero shot Datasets}} & \multicolumn{2} {c} {\textbf{Statistics}}\\
         \cmidrule(l{0.6em}r{0.6em}){2-17} \cmidrule(l{0.6em}r{0.6em}){18-19}
        & \multicolumn{2} {c|} {\textbf{CalTech101}} & \multicolumn{2} {c|} {\textbf{Cars}} & \multicolumn{2} {c|} {\textbf{DTD}} & \multicolumn{2} {c|} {\textbf{EuroSAT}} & \multicolumn{2} {c|} {\textbf{FGVC}} & \multicolumn{2} {c|} {\textbf{Flowers}} & \multicolumn{2} {c|} {\textbf{Oxford Pets}} & \multicolumn{2} {c|} {\textbf{STL-10}} & \multicolumn{2} {c} {\textbf{ZS Avg.}} \\
        Method & Clean & Adv & Clean & Adv & Clean & Adv & Clean & Adv & Clean & Adv & Clean & Adv & Clean & Adv & Clean & Adv & Clean & Adv \\
    \toprule
    \midrule
     CLIP & 80.47 & 0.00 & 51.47 & 0.00 & 41.43 & 0.00 & 45.27 & 0.00 & 15.93 & 0.00 & 60.18 & 0.00 & 84.00 & 0.00 & 95.71 & 0.00 & 59.31 & 0.00 \\
     \cmidrule{1-19}\noalign{\vskip 0.5ex}
      Vanilla FT & 79.85 & 0.00 & 45.44 & 0.00 & 40.26 & 0.00 & 41.18 & 0.00 & 14.16 & 0.00 & 55.42 & 0.00 & 84.30 & 0.00 & 97.03 & 0.00 & 57.21 & 0.00 \\
      WISE-FT & 81.47 & 0.00 & 53.18 & 0.00 & 42.07 & 0.00 & 46.57 & 0.00 & 16.08 & 0.00 & 60.66 & 0.00 & 85.90 & 0.00 & 97.26 & 0.00 & \textbf{60.40} & 0.00 \\
      FLYP & 80.77 & 0.00 & 33.72 & 0.00 & 38.56 & 0.00 & 39.72 & 0.00 & 11.76 & 0.00 & 50.18 & 0.00 & 84.21 & 0.00 & 96.00 & 0.00 & 54.37 & 0.00 \\
      TPGM & 80.49 & 0.00 & 47.69 & 0.00 & 40.85 & 0.00 & 43.37 & 0.00 & 13.86 & 0.00 & 57.18 & 0.00 & 85.14 & 0.00 & 97.25 & 0.00 & 58.23 & 0.00 \\
      SPD & 79.80 & 0.00 & 46.31 & 0.00 & 40.53 & 0.00 & 42.51 & 0.00 & 15.03 & 0.00 & 57.70 & 0.00 & 84.84 & 0.00 & 97.38 & 0.00 & 58.01 & 0.00 \\
      \cmidrule{1-19}\noalign{\vskip 0.5ex}
      TeCoA & 74.03 & 51.40 & 10.68 & 1.50 & 24.04 & 10.70 & 18.72 & 11.00 & 4.56 & 0.50 & 25.59 & 6.60 & 67.07 & 29.10 & 85.30 & 55.40 & 38.75 & 20.78 \\
      FARE & 76.97 & 45.80 & 33.87 & 1.30 & 27.87 & 12.20 & 14.77 & 11.00 & 8.73 & 0.60 & 31.72 & 5.30 & 75.44 & 20.40 & 87.87 & 55.50 & 44.66 & 19.01 \\
      PMG-AFT & 77.10 & 53.20 & 29.50 & 2.00 & 30.20 & 12.90 & 22.30 & 11.50 & 7.90 & 0.80 & 33.40 & 7.30 & 72.10 & 31.50 & 88.90 & 56.80 & 45.18 & \textbf{22.00} \\
      LAAT & 75.10 & 47.20 & 28.80 & 1.60 & 29.90 & 11.80 & 20.60 & 10.90 & 6.10 & 0.70 & 28.90 & 6.10 & 72.20 & 22.40 & 86.10 & 55.90 & 43.46 & 19.58 \\
      \midrule
      \rowcolor{gray!15}
      GRACE & 79.23 & 49.30 & 52.10 & 1.90 & 41.60 & 12.60 & 46.10 & 11.20 & 15.50 & 0.70 & 60.05 & 7.00 & 85.20 & 30.10 & 97.10 & 56.00 & \underline{59.61} & \underline{21.10} \\
    \bottomrule
    \end{tabular}
    }
    \vspace{-10pt}
    \caption{\textbf{Clean and adversarial evaluation on zero-shot image classification datasets (ViT-B/32).} Models are trained on ImageNet; all other datasets are zero-shot. ZS Avg is the mean across the 8 datasets.\vspace{-15pt}}
    \label{tab:zero_shot_vit_b_32}
\end{table*}

\begin{table}[t]
\centering
\small
\resizebox{0.8\linewidth}{!}{
\begin{tabular}{c|c|c|c|c}
        \toprule
        Method & ID & Average & Average & Harmonic \\
         & & OOD & Adversarial & Mean \\
         \toprule
        \midrule
        CLIP & 63.35 & 57.44 & 8.82 & 20.46 \\
        \midrule
        Vanilla FT & 74.86 & 56.59 & 8.95 & 21.01 \\
        WISE-FT & 70.20 & 58.05 & 9.04 & 21.11 \\
        FLYP & 72.15 & 55.32 & 8.49 & 20.03 \\
        TPGM & 74.13 & 57.56 & 8.89 & 20.92 \\
        SPD & 73.68 & 57.53 & 8.69 & 20.54 \\
        \midrule
        TeCoA & 52.54 & 37.96 & 17.48 & 29.24 \\
        FARE & 47.38 & 41.56 & 13.43 & 25.07 \\
        PMG-AFT & 58.20 & 43.40 & 19.57 & \underline{32.85} \\
        LAAT & 55.46 & 41.95 & 17.90 & 30.69 \\
        \midrule
        \rowcolor{gray!15}
        GRACE (Ours) & 74.21 & 57.01 & 22.44 & \textbf{39.69} \\
        \bottomrule
    \end{tabular}
}
\caption{\textbf{Unified summary across settings (ViT-B/32).} CLIP ViT-B/32 fine-tuned on ImageNet-1K and evaluated across ID, OOD, PGD adversarial (AutoAttack/APGD-CE), and natural adversarial (ImageNet-A/A-Plus average). \vspace{-20pt}}
\label{tab:results}
\end{table}

\section{Experiments and Results}
\label{sec:experiments}

\subsection{Problem Setting}
\label{sec:problem_setting}
\textbf{Testing OOD Robustness.} We evaluate the out-of-distribution (OOD) robustness of fine-tuned Vision–Language Models on image classification tasks. Our experiments employ Vision Transformers (ViTs)~\cite{transformer} pre-trained with CLIP~\cite{clip}, using the ViT-B/32 backbone for the main study. The model is fine-tuned on ImageNet-1K and tested on three OOD benchmarks: ImageNet-V2~\cite{imagenetv2}, ImageNet-R~\cite{imagenet-r}, ImageNet-S~\cite{imagenet-s}. No OOD data is used during training. To further assess zero-shot generalization, we evaluate on eight datasets covering diverse visual domains—Caltech101, Cars, DTD, EuroSAT, FGVC, Flowers102, Oxford Pets, and STL-10—following standard CLIP zero-shot protocols.

\textbf{Testing Adversarial Robustness.} We evaluate robustness to both natural and targeted adversarial robustness in our experiments. For testing robustness to natural adversarial data, we use the ImageNet-A~\cite{imagenet-a} and ImageNet-A-Plus~\cite{imagenet-a-plus} datasets. During training, adversarial samples are generated using 10-step PGD under an $\ell_\infty$ perturbation radius of $4/255$ and step size $1/255$.  At test time, we employ AutoAttack~\cite{autoattack} (APGD-CE) at the same perturbation radius, ensuring consistent, attack-agnostic evaluation.  
We report both clean and adversarial accuracies across ID, OOD, and zero-shot settings.  

\textbf{Baselines.} We compare GRACE with (i) the standard fine-tuning baseline,  (ii) \emph{robust fine-tuning} methods—WiSE-FT~\cite{wiseft}, FLYP~\cite{flyp}, TPGM~\cite{tpgm}, and SPD~\cite{spd},  
and (iii) \emph{adversarial fine-tuning} methods—TeCoA~\cite{tecoa}, FARE~\cite{fare}, PMG-AFT~\cite{pmg_aft}, and LAAT~\cite{laat}.  
All baselines use the same CLIP prompts, preprocessing, and optimization schedule for fair comparison.

\subsection{Results}
\label{sec:results}
\paragraph{Unified Comparison across Settings.}
Table~\ref{tab:results} presents the unified performance summary across \textbf{ID}, \textbf{OOD}, \textbf{adversarial}, and \textbf{zero-shot} regimes for the ViT-B/32 backbone.  
GRACE consistently improves robust performance across all domains, achieving an average +13.6\% increase in PGD adversarial accuracy while and maintaining natural adversarial performance compared to vanilla fine-tuning.  
Notably, GRACE maintains high ID accuracy (74.2\%), only marginally below the vanilla baseline, while achieving large gains in both distribution-shift and adversarial conditions—indicating effective flattening of the loss landscape without sacrificing generalization.  
Among all methods, GRACE strikes the most balanced trade-off between ID, OOD, and adversarial robustness.

\begin{table}[t]
\centering
\small
\resizebox{0.7\linewidth}{!}{
\begin{tabular}{c|c|c|c|c}
        \toprule
        Method & ID & Average & Average & Harmonic \\
         & & OOD & Adversarial & Mean \\
         \toprule
        \midrule
        CLIP & 63.3 & 57.4 & 8.8 & 20.4 \\
        \midrule
        LoRA-FT & 72.8 & 55.0 & 8.2 & 19.4 \\
        LoRA-SPD & 73.0 & 56.0 & 8.5 & 19.5 \\
        LoRA-TeCoA & 60.0 & 45.0 & 22.5 & 37.5 \\
        VPT-PMG-AFT & 70.0 & 52.0 & 22.7 & 38.6 \\
        \midrule
        \rowcolor{gray!15}
        GRACE & 74.2 & 57.0 & 22.4 & 39.6 \\
        \bottomrule
    \end{tabular}
}
\caption{\textbf{Comparison of \emph{GRACE} with other LoRA-based fine-tuning approaches.} CLIP ViT-B/32 fine-tuned on ImageNet-1K. \vspace{-15pt}}
\label{tab:peft}
\end{table}

\textbf{Out-of-Distribution Robustness.} The detailed OOD performance is reported in Table~\ref{tab:imagenet_vit_b_32}. Compared to strong OOD-oriented baselines like WiSE-FT and SPD, GRACE improves maintains the average OOD and ID accuracy improving Adversari accuracy by 13\%. TeCoA and FARE, despite their strong adversarial robustness, underperform on ImageNet-V2/S/R due to overfitting to adversarial distributions.  In contrast, GRACE preserves pre-trained structure via curvature-aware regularization,enabling robustness transfer across all domains. Zero-shot evaluations in Table \ref{tab:zero_shot_vit_b_32} (bottom) further confirm that GRACE maintains competitive clean performance while improving adversarial accuracy by +2.09\% over FARE.

\textbf{Adversarial Robustness.} GRACE substantially outperforms prior adversarial fine-tuning methods under both PGD and natural adversarial conditions. As shown in Table~\ref{tab:results}, GRACE achieves the highest adversarial accuracy across ID, OOD, and ZS settings (+2.87\% over PMG-AFT on average) while preserving OOD stability. This demonstrates that layerwise adaptive rank perturbations effectively flatten the anisotropic curvature of the weight-loss landscape, preventing robustness–generalization collapse.

\textbf{Findings.} Overall, GRACE bridges the three-way trade-off among ID accuracy, OOD generalization, and adversarial robustness. Unlike prior methods that specialize in one axis (e.g., WiSE-FT for OOD, ~TeCoA for adversarial), GRACE jointly optimizes for all through curvature-aware, low-rank perturbation and Gram-based invariance alignment.  
We observe consistent improvements in harmonic mean average across settings, as discussed in Section~\ref{sec:ablations}.

\subsection{Comparison with Other LoRA-based PEFT Approaches}
\label{sec:peft}
We further benchmark GRACE against other LoRA-style parameter-efficient fine-tuning methods in Table~\ref{tab:peft}.  
Compared to LoRA-FT, LoRA-SPD, and LoRA-TeCoA, GRACE achieves the best overall performance (39.6\%), highlighting the benefits of adaptive perturbation over static low-rank updates. These results emphasize that robustness-oriented PEFT requires dynamic control of perturbation rank and curvature sensitivity, as implemented in GRACE.

\vspace{6pt}
\subsection{Ablation Study}
\label{sec:ablations}
To assess the contribution of each component in GRACE, we perform ablations on the ViT-B/32 model fine-tuned on ImageNet (Table~\ref{tab:ablations}).  
Adding the \emph{Gram-volume (GV)} term improves OOD generalization (+1.5\%), while including \emph{LAR-AWP} without a curriculum primarily enhances adversarial robustness (+8.6\%). The full \textbf{GRACE} model, combining both modules with a rank-adaptive curriculum, achieves the best average, demonstrating the complementary effects of representation consistency and curvature regularization. These results confirm that both geometric components are critical for stabilizing representations and mitigating catastrophic forgetting during fine-tuning.

\begin{table}[t]
\centering
\small
\label{tab:ablations}
\resizebox{0.85\linewidth}{!}{
\begin{tabular}{c|c|c|c|c}
        \toprule
        Variant & ID & Average & Average & Harmonic \\
         & & OOD & Adversarial & Mean \\
         \toprule
        \midrule
        CLIP & 63.3 & 57.4 & 8.8 & 20.4 \\
        \midrule
        LoRA-FT (no regs) & 72.8 & 55.0 & 8.2 & 19.4 \\
        + GV only & 72.0 & 56.5 & 8.6 & 20.2 \\
        + LAR-AWP (no curriculum) & 71.0 & 53.0 & 17.2 & 32.9 \\
        + LAR-AWP (with curriculum) & 72.5 & 54.0 & 22.2 & 38.7 \\
        \midrule
        \rowcolor{gray!15}
        GRACE & 74.2 & 57.0 & 22.4 & 39.6 \\
        \bottomrule
    \end{tabular}
}
\caption{\textbf{Ablation Study.} CLIP ViT-B/32 fine-tuned on ImageNet-1K. \emph{Avg} is the simple mean of the four columns.\vspace{-15pt}}
\label{tab:ablations}
\end{table}

\begin{figure}[t]
    \centering
    \includegraphics[width=0.77\linewidth]{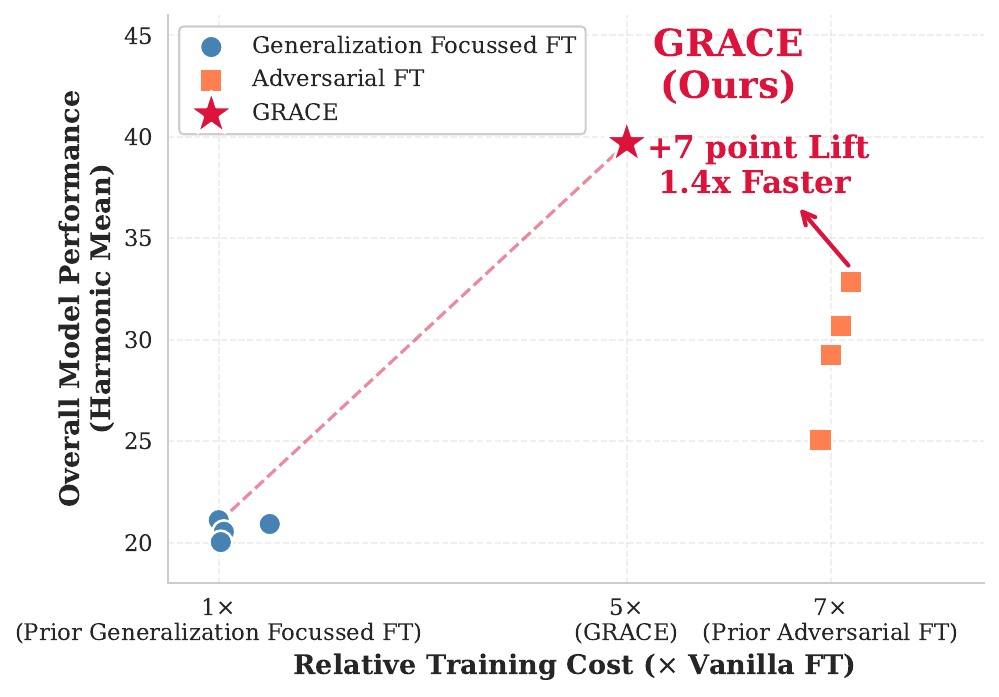}
    \vspace{-9pt}
    \caption{\textbf{Pareto Curve.} GRACE (Red) achieves +7 performance gain while being 1.4$\times$ faster than prior adversarial methods.\vspace{-15pt}}
    \label{fig:pareto}
\end{figure}

\subsection{Computation Time Analysis}
In this section, we analyze the computational efficiency of the proposed GRACE framework and compare it to vanilla fine-tuning and adversarial fine-tuning approaches. As shown in the Pareto Frontier (Fig. \ref{fig:pareto}), GRACE provides a more favorable robustness–compute trade-off than prior adversarial FT methods (e.g., TeCoA, PMG-AFT), achieving higher overall performance at a lower training cost.

%% file: sec/7_conclusion.tex
\vspace{-5pt}
\section{Conclusion}
\label{sec:conclusion}
\vspace{-5pt}
In this work, we present \textbf{GRACE}—\textit{Gram-aligned Robustness via Adaptive Curvature Estimation}—a unified framework for fine-tuning VLMs that bridges the gap between ID performance, OOD generalization, and adversarial robustness. Grounded in a robust PAC-Bayesian analysis, GRACE jointly optimizes parameter-space flatness and feature-space invariance through two complementary components: Layer-wise Adaptive Low-Rank Adversarial Weight Perturbation and Gram-Volume Alignment. Our geometric diagnostics reveal that GRACE produces flatter loss landscapes and more stable feature manifolds, leading to consistent gains across ID, OOD, and adversarial regimes on ImageNet benchmarks. We hope this work establishes geometry-driven fine-tuning as a general paradigm for robust multimodal adaptation.

\section{Acknowledgement}
This material is based upon work partially supported by the National Science Foundation under Grant No. 2239292.

%% file: sec/X_suppl.tex
\clearpage
\setcounter{page}{1}
\maketitlesupplementary

\appendix

\section{Layerwise Curvature Analysis of CLIP}
\label{app:layerwise-curvature}

To examine the geometric heterogeneity of Vision–Language Models, we conduct a \emph{layerwise Hessian curvature analysis} of the CLIP ViT models.  
Whereas global curvature metrics (e.g., top eigenvalue or Frobenius norm of the full Hessian) summarize the overall sharpness of the loss landscape, they obscure the internal variation across transformer depth.  
Our goal is to resolve this structure by measuring second-order curvature within each residual block.

\begin{figure}[h]
    \centering
    \includegraphics[width=\linewidth]{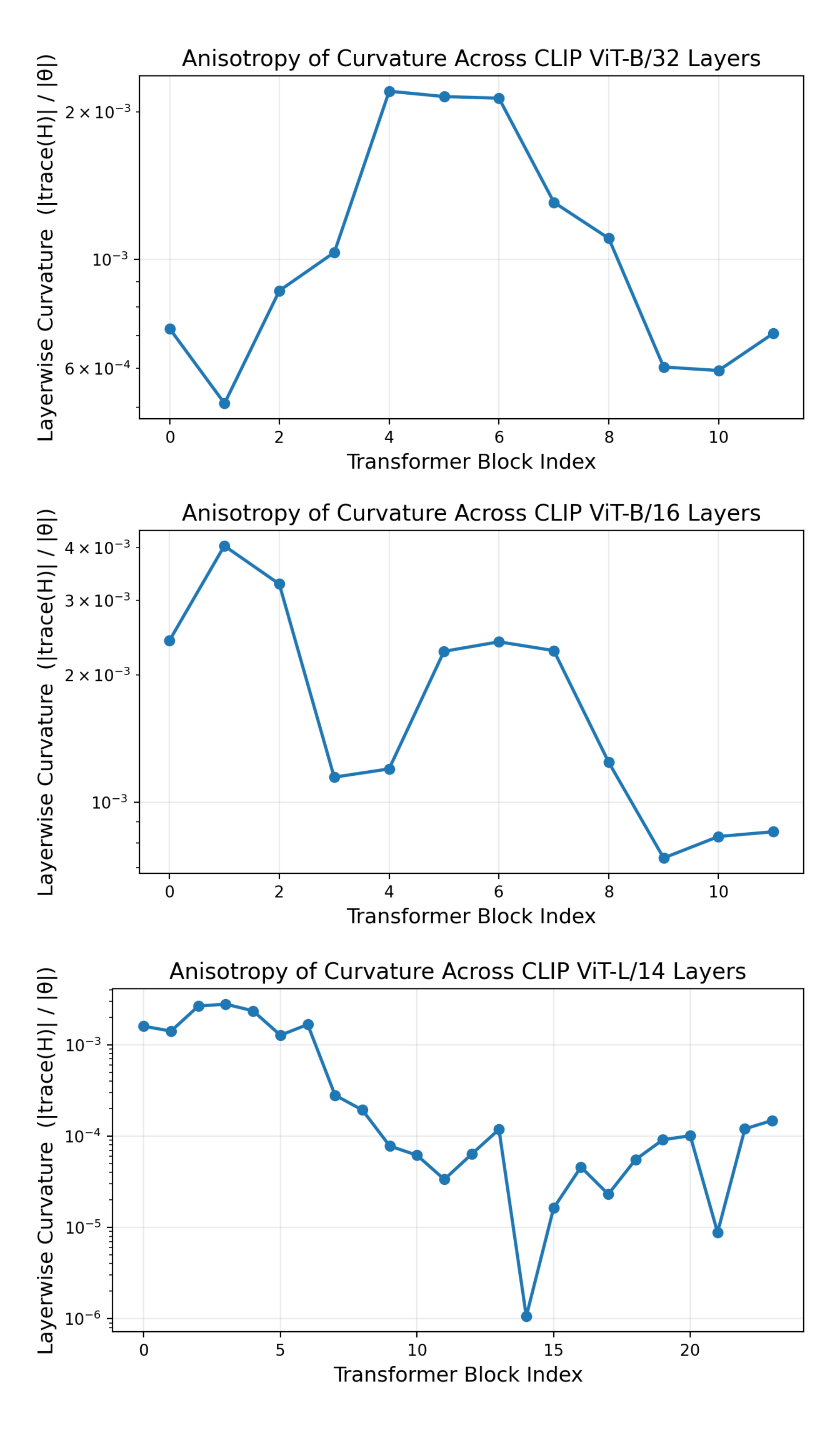}
    \caption{\textbf{Layerwise curvature anisotropy in CLIP.}
    Normalized Hutchinson curvature $\kappa_\ell$ for each transformer block of CLIP ViT-B/32, ViT-B/16, and ViT-L/14. All models exhibit substantial variation in curvature across depth, indicating strong layerwise geometric heterogeneity.}
    \label{fig:layerwise-curvature}
\end{figure}

\paragraph{Method.}
Let the model parameters be partitioned into disjoint blocks $\{\theta_\ell\}$ corresponding to the $\ell$-th transformer layer.  
For a batch loss $\mathcal{L}(\theta)$, we estimate the trace of the blockwise Hessian
$H_\ell = \nabla_{\theta_\ell}^2 \mathcal{L}$
using a block-restricted Hutchinson estimator:
\begin{equation}
\label{eq:hutch-layer}
\operatorname{tr}(H_\ell)
\;\approx\;
\frac{1}{m}
\sum_{j=1}^m
\big(v_{\ell}^{(j)}\big)^{\!\top}
\big(H v^{(j)}\big),
\end{equation}
where $v^{(j)} \sim \mathcal{N}(0,I)$ and $v_{\ell}^{(j)}$ is the subvector associated with block $\ell$.  
We use $m{=}750$ probe vectors and compute Hessian–vector products via double backpropagation.  
We report the normalized curvature
\begin{equation}
\label{eq:norm-layer-curv}
\kappa_\ell
= \frac{\operatorname{tr}(H_\ell)}{\|\theta_\ell\|_0},
\end{equation}
which approximates the average curvature per parameter in block~$\ell$.

\paragraph{Findings.}
As shown in Fig.~\ref{fig:layerwise-curvature}, the curvature $\kappa_\ell$ varies substantially across depth for all three CLIP variants (ViT-B/32, ViT-B/16, and ViT-L/14).  
Some layers operate in relatively sharp regions of the loss landscape, while others lie in significantly flatter regimes.  
This non-uniformity demonstrates that CLIP exhibits strong \emph{layerwise anisotropy} in its second-order geometry.

\paragraph{Implications.}
Because curvature differs by more than an order of magnitude across layers, applying a single perturbation strength or smoothing radius---as in uniform SAM, AWP, or adversarial fine-tuning---is inherently misaligned with the model’s structure.  
This motivates our \emph{Layerwise Adaptive Adversarial Weight Perturbation}, which modulates perturbation rank based on local curvature estimates to better match the geometry of each layer.

\section{Preliminaries}
\label{app:preliminaries-extra}

\subsection{Image Classification with Vision--Language Models}
\label{sec:ood_setup}

In a $K$-class image classification problem with inputs $x \in \mathcal{X}$ and labels $y \in \{1,\dots,K\}$,
CLIP-style VLMs frame classification as image--text matching~\cite{clip}.
Each class label $k$ is converted into a caption using a prompt template such as
``\texttt{A photo of a \{label\}}'',
yielding prompts $\{t_k\}_{k=1}^K$.

The text encoder $G_\phi$ produces class-specific text embeddings
$g_k = G_\phi(t_k) \in \mathbb{R}^D$,
which are stacked to form the classification head
\begin{equation}
    \mathbf{W}
    =
    \begin{bmatrix}
    G_\phi(t_1)^\top \\
    \vdots \\
    G_\phi(t_K)^\top
    \end{bmatrix}
    \in \mathbb{R}^{K\times D}.
\end{equation}
The image encoder $F_\theta$ maps an input image $x$ to an embedding $z(x)=F_\theta(x)\in\mathbb{R}^D$
(often $\ell_2$-normalized).
Class logits for $x$ are then
\begin{equation}
    u(x) = \mathbf{W} z(x) \in \mathbb{R}^K,
\end{equation}
and the predictive distribution is given by the softmax
\begin{equation}
    p(y=k \mid x) = \frac{\exp(u^{(k)}(x))}{\sum_{j=1}^{K} \exp(u^{(j)}(x))},
    \qquad k=1,\dots,K.
\end{equation}

\subsection{Fine-Tuning VLMs for Image Classification}
\label{sec:finetuning}

Although VLMs exhibit strong zero-shot performance across diverse domains,
their accuracy on a specific downstream dataset $\mathcal{D}=\{(x,y)\}$
can typically be improved via fine-tuning.
A standard approach is to update all or some subset of the model parameters $\theta$ to minimize the cross-entropy loss
\begin{equation}
    \mathcal{L}_{\text{CE}}(\theta,\mathbf{W})
    =
    -\sum_{(x,k)\in\mathcal{D}}
    \log\big(\text{Softmax}^{(k)}(\mathbf{W}F_\theta(x))\big).
\end{equation}
Full fine-tuning (updating all of $\theta$) can yield strong in-distribution performance,
but is computationally costly and prone to overfitting and catastrophic forgetting.
This motivates parameter-efficient adaptation schemes such as LoRA.

\subsection{Low-Rank Adaptation (LoRA)}
\label{app:lora-details}

\begin{figure}[t]
    \centering
    \includegraphics[width=0.8\linewidth]{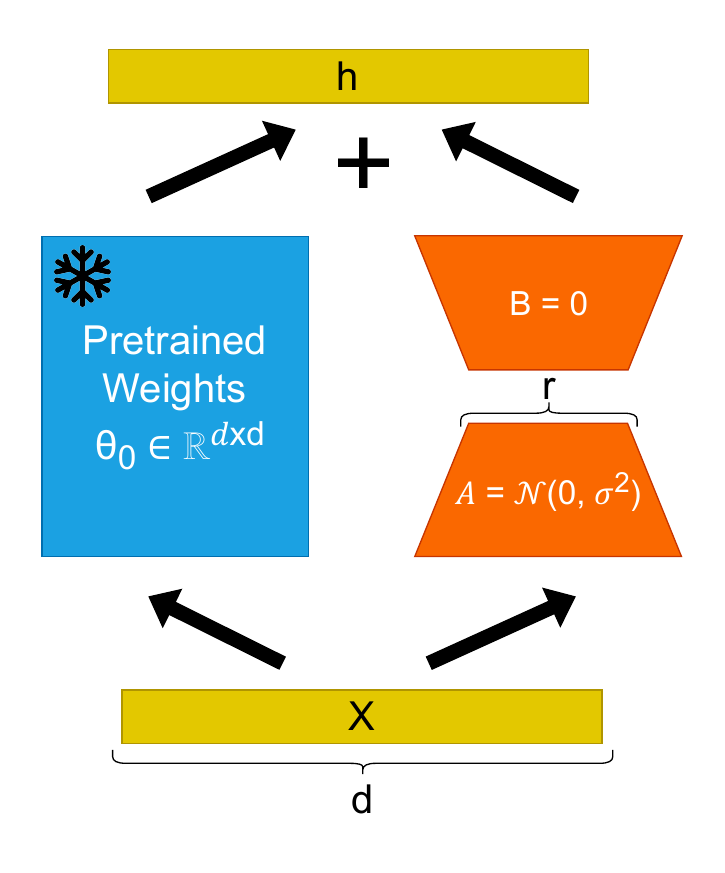}
    \caption{\textbf{Schematic of LoRA weight updates.}
    Weight updates are parameterized by low-rank matrices $A$ and $B$ of rank $r$,
    constraining deviations from the pre-trained weights to a low-dimensional subspace.}
    \label{fig:lora}
\end{figure}

Large-scale VLMs are heavily over-parameterized, which makes full fine-tuning expensive
and increases the risk of drifting away from the pre-trained manifold.
Low-Rank Adaptation (LoRA)~\cite{lora} mitigates this by restricting weight updates
to a low-rank subspace.
For any weight matrix $W \in \mathbb{R}^{n_1 \times n_2}$ in the pre-trained model,
LoRA parameterizes its fine-tuned version as
\begin{equation}
    W = W^* + \frac{\alpha}{r} B A,
\end{equation}
where $W^*$ denotes the frozen pre-trained weights,
and $B \in \mathbb{R}^{n_1 \times r}$, $A \in \mathbb{R}^{r \times n_2}$ are trainable low-rank factors.
The rank $r \ll \min(n_1,n_2)$ and scaling factor $\alpha \in \mathbb{R}$ are hyperparameters.
Only $A$ and $B$ are updated; $W^*$ remains fixed.
This reduces the number of trainable parameters,
improves efficiency, and implicitly constrains the adaptation to remain close to the pre-trained solution,
consistent with the KL term in our PAC-Bayesian analysis. The overall schematics of the low-rank weight updates in LoRA can be seen in Figure \ref{fig:lora}.

\subsection{Adversarial Training}
\label{app:adv-training-details}

Adversarial training~\cite{madry2018towards} aims to make neural networks robust
to perturbations within a bounded norm ball.
Given a model $f$ and a perturbation set
$\mathcal{S} = \{\delta: \|\delta\|_p \le \epsilon\}$,
the robust optimization objective is
\begin{equation}
    \min_{f}
    \mathbb{E}_{(x,y)\sim\mathcal{D}}
    \Big[
      \max_{\delta \in \mathcal{S}}
      \mathcal{L}(f(x+\delta),y)
    \Big].
\end{equation}
The inner maximization can be approximated by gradient-based methods:

\paragraph{Fast Gradient Sign Method (FGSM).}
FGSM performs a single-step update
\begin{equation}
    x_{\text{adv}}
    =
    x + \epsilon \,\mathrm{sgn}\big(\nabla_x \mathcal{L}(f(x),y)\big),
\end{equation}
where $\mathrm{sgn}(\cdot)$ is the element-wise sign function.

\paragraph{Projected Gradient Descent (PGD).}
PGD performs a multi-step refinement
\begin{equation}
    x^{t+1}
    =
    \Pi_{\mathcal{S}}\Big(x^t + \gamma\, \mathrm{sgn}(\nabla_{x^t}\mathcal{L}(f(x^t),y))\Big),
\end{equation}
where $\gamma$ is the step size, $t$ indexes the iteration,
and $\Pi_{\mathcal{S}}$ is the projection operator onto the perturbation set $\mathcal{S}$.
PGD-based adversarial training is the standard baseline for robust learning,
but often converges to sharp minima and can harm natural generalization~\cite{landscape,awp}.

\subsection{Adversarial Weight Perturbation (AWP)}
\label{app:awp-details}

The geometry of the weight--loss landscape is strongly correlated
with both standard and robust generalization~\cite{visualloss,landscape}.
Adversarial Weight Perturbation (AWP)~\cite{awp} introduces a second adversary in parameter space
to explicitly regularize loss-landscape flatness under adversarial training.

Let $\theta$ denote the current model parameters and $\mu$ an adversarial weight perturbation.
AWP alternates between adversarial input and weight updates:

\paragraph{Adversarial input update.}
Starting from $x^0 = x$, and with $\mu$ initially set to $0$,
AWP first generates adversarial examples using the perturbed model:
\begin{equation}
   x^{t+1}
   =
   \Pi_{\mathcal{S}}
   \Big(
      x^t + \gamma_1\, \mathrm{sgn}\big(\nabla_{x^t}\mathcal{L}(F_{\theta+\mu}(x^t),y)\big)
   \Big),
\end{equation}
for a chosen number of steps and step size $\gamma_1$.

\paragraph{Adversarial weight update.}
Given the adversarial samples $\{x_i^t\}_{i=1}^m$,
AWP then updates the weight perturbation $\mu$ by ascending the loss:
\begin{equation}
    \mu
    =
    \Pi_{\Gamma}
    \Big(
      \mu
      + \gamma_2
      \frac{
        \nabla_\mu \frac{1}{m}
        \sum_{i=1}^m \mathcal{L}(F_{\theta+\mu}(x_i^t),y_i)
      }{
        \big\|
        \nabla_\mu \frac{1}{m}
        \sum_{i=1}^m \mathcal{L}(F_{\theta+\mu}(x_i^t),y_i)
        \big\|_2
      }
      \,\|\theta\|
    \Big),
\end{equation}
where $\Pi_{\Gamma}$ projects onto a bounded perturbation set in parameter space,
and $\gamma_2$ is a step size.
This update can be done in one or multiple steps.

\paragraph{Model parameter update.}
Finally, the base parameters are updated via SGD on the perturbed model:
\begin{equation}
    \theta
    =
    (\theta + \mu)
    - \gamma_3 \nabla_{\theta+\mu}
      \frac{1}{m}\sum_{i=1}^m \mathcal{L}(F_{\theta+\mu}(x_i^t),y_i)
    - \mu,
\end{equation}
with learning rate $\gamma_3$.
Intuitively, $\mu$ identifies directions in parameter space where the adversarial loss increases sharply,
and the outer update drives the model toward flatter, more robust regions.

\section{Extended Theoretical Analysis}
\label{app:etended_theory}

This section provides complete proofs and extended derivations for the theoretical results presented in Section~\ref{sec:theory}. 
We include proofs for Theorem~\ref{thm:main} (Robust PAC-Bayes Bound) 
and Lemma~\ref{lem:feature}, additional supporting lemmas, 
and clarifications on the assumptions used in the main text.

\subsection{Notation}

Let $\theta_0$ denote the pre-trained parameters of a VLM encoder, 
and $\theta$ its fine-tuned counterpart. 
The risk for domain $s \in \{ \mathrm{ID}, \mathrm{OOD}, \mathrm{Adv} \}$ is
\[
R_s(\theta)
= \mathbb{E}_{(x,y) \sim \mathcal{D}_s}[\ell(f_\theta(x), y)].
\]
The robust risk is $R_{\mathrm{Rob}}(\theta) = \max_{s} R_s(\theta)$.
We denote by $P = \mathcal{N}(\theta_0, \sigma^2 I)$ the prior
and $Q = \mathcal{N}(\theta, \sigma^2 I)$ the posterior.

\subsection{Proof of Theorem~\ref{thm:main}}
\label{app:proof-main}

We now provide the complete derivation of the robust PAC-Bayesian bound.

\subsubsection{PAC-Bayes Preliminaries}

We use the standard PAC-Bayesian inequality  
\cite{mcallester2013pacbayesiantutorialdropoutbound,neyshabur2018pacbayesianapproachspectrallynormalizedmargin}:
for any prior distribution $P$ and posterior $Q$ over model parameters,
with probability at least $1-\delta$ over the training set,
\begin{equation}
\mathbb{E}_{\theta' \sim Q}[R(\theta')]
\leq
\mathbb{E}_{\theta'\sim Q}[\hat{R}(\theta')]
+ \sqrt{\frac{ KL(Q\|P) + \ln(2\sqrt{n}/\delta)}{2n} }.
\label{eq:pac-prelim}
\end{equation}

To extend this to the robust risk 
$R_{\mathrm{Rob}}(\theta) = \max_s R_s(\theta)$, 
we apply a union bound over the domains.

\subsubsection{KL Divergence under Low-Rank Adaptation (LoRA)}
\label{app:lora-kl}

In the main text we consider a Gaussian prior 
$P = \mathcal{N}(\theta_0,\sigma^2 I)$ 
and posterior 
$Q = \mathcal{N}(\theta,\sigma^2 I)$.
When fine-tuning with LoRA, however, only a low-dimensional subspace 
of the parameters is updated.  
Let the LoRA update be 
$\Delta\theta = W_{\mathrm{LoRA}} = BA$, 
where 
$A \in \mathbb{R}^{r\times k}$ and 
$B \in \mathbb{R}^{d\times r}$,
with $r \ll d$.
Thus the effective parameterization lies in an $rk$-dimensional subspace.

We therefore define the prior and posterior over the LoRA parameters:
\[
P = \mathcal{N}(0, \sigma^2 I_{rk}), \qquad
Q = \mathcal{N}\big((A,B), \sigma^2 I_{rk}\big).
\]

The KL divergence becomes:
\begin{equation}
KL(Q\|P)
= \frac{1}{2\sigma^2}
\big( \|A\|_F^2 + \|B\|_F^2 \big)
= \frac{\|W_{\mathrm{LoRA}}\|_F^2}{2\sigma^2}.
\label{eq:lora-kl}
\end{equation}

LoRA therefore reduces the complexity term in the PAC-Bayesian bound 
by restricting the posterior to a low-rank subspace.  
This provides a natural proximity regularization mechanism, ensuring 
that fine-tuning remains close to the pre-trained parameters 
while controlling the variance of the posterior.

\subsubsection{Second-Order Expansion and Sharpness Term}

Using Assumption~\ref{asmp:smooth} (bounded Hessian),
we apply a second-order Taylor expansion around $\theta$:
\[
\ell(\theta + \epsilon)
= \ell(\theta)
+ \epsilon^\top \nabla\ell(\theta)
+ \frac{1}{2} \epsilon^\top \nabla^2\ell(\theta) \epsilon
+ O(\|\epsilon\|^3).
\]

Taking expectation w.r.t. $\epsilon \sim \mathcal{N}(0,\sigma^2 I)$:
\begin{align}
\mathbb{E}_{\epsilon}[\ell(\theta + \epsilon)]
&= \ell(\theta)
+ \frac{\sigma^2}{2} \operatorname{Tr}(\nabla^2_\theta \ell(\theta))
+ O(\sigma^3).
\end{align}

Applying the same to $R_{\mathrm{Rob}}$ yields:
\begin{equation}
\mathbb{E}_{\theta'\sim Q}[R_{\mathrm{Rob}}(\theta')]
=
R_{\mathrm{Rob}}(\theta)
+ \frac{\sigma^2}{2} \operatorname{Tr}(\nabla^2_\theta R_{\mathrm{Rob}}(\theta))
+ O(\sigma^3).
\label{eq:sharpness-term}
\end{equation}

The trace term acts as an average curvature (``sharpness'') penalty.

\subsubsection{Union Bound for Multi-Domain Robust Risk}

The robust risk satisfies:
\[
R_{\mathrm{Rob}}(\theta) = \max_s R_s(\theta)
\leq
\sum_s R_s(\theta).
\]
Applying PAC-Bayes to each domain and union bounding over 
$|\mathcal{S}|=3$ domains gives a factor $\ln(2n/\delta)$.

\subsubsection{Final Bound}

Combining \eqref{eq:pac-prelim}, \eqref{eq:lora-kl}, \eqref{eq:sharpness-term} 
for the robust case yields:
\begin{dmath}
R_{\mathrm{Rob}}(\theta)
\leq
\hat{R}_{\mathrm{ID}}(\theta)
+ \frac{\|W_{\mathrm{LoRA}}\|_F^2}{2\sigma^2}
+ \frac{\sigma^2}{2} \operatorname{Tr}(\nabla^2 R_{\mathrm{Rob}}(\theta))
+ \max_{s,t} d_{\mathcal{H}\Delta\mathcal{H}}(\mathcal{D}_s,\mathcal{D}_t)
+ \lambda^*,
\end{dmath}
which completes the proof.

\hfill $\square$

\subsection{Proof of Lemma~\ref{lem:feature}}
\label{app:proof-lemma}

We derive an upper bound on the $\mathcal{H}\Delta\mathcal{H}$-divergence
between domains $s$ and $t$ under a Lipschitz encoder.

\subsubsection{Feature-Space Decomposition}

The $\mathcal{H}\Delta\mathcal{H}$-divergence satisfies
\cite{NIPS2006_b1b0432c}:
\[
d_{\mathcal{H}\Delta\mathcal{H}}(\mathcal{D}_s,\mathcal{D}_t)
= 2 \sup_{h,h'\in\mathcal{H}}
\Big|
\Pr_s[h\neq h'] - \Pr_t[h\neq h']
\Big|.
\]

For linear separators on a feature map $\phi(x)$,
this reduces to the discrepancy between distributions of $\phi(x)$.
Using Assumption~\ref{asmp:features}
(Lipschitz continuity of $f_\theta$),
\[
\|f_\theta(x) - f_\theta(x')\|
\leq L_f \|\theta - \theta'\|.
\]

Let $\mu_s^c,\Sigma_s^c$ denote class-conditional mean and covariance.
A standard result on Wasserstein/TV bounds yields:
\[
d(\mathcal{D}_s,\mathcal{D}_t)
\leq 
2 L_f 
\sum_{c=1}^k \pi_c
\left(
 \|\mu_s^c - \mu_t^c\|_2
 + \|\Sigma_s^c - \Sigma_t^c\|_F
\right),
\]
and noting $\|A\|_F = \sqrt{\operatorname{Tr}(A^\top A)}$ produces
the stated bound:
\[
d_{\mathcal{H}\Delta\mathcal{H}}(\mathcal{D}_s,\mathcal{D}_t)
\leq 
2L_f \sum_{c=1}^k \pi_c 
\left(
\|\mu_s^c - \mu_t^c\|_2
+ \sqrt{
\operatorname{Tr}(\Sigma_s^c - \Sigma_t^c)^2
}
\right).
\]

\hfill $\square$

\subsection{Supporting Lemmas and Clarifications}

\begin{lemma}[Hessian Boundedness]
If $\|\nabla^2 \ell\| \leq M$ for all samples, then
\[
\operatorname{Tr}(\nabla^2 R(\theta))
= \mathbb{E}[\operatorname{Tr}(\nabla^2 \ell(\theta))]
\leq M d.
\]
\end{lemma}

\begin{lemma}[Taylor Remainder]
Under Assumption~\ref{asmp:smooth}, the third-order term satisfies
\[
|R_3| \leq \frac{M}{6} \mathbb{E}\|\epsilon\|^3 = O(\sigma^3).
\]
\end{lemma}

\subsubsection{Discussion of Assumptions}

\paragraph{Smoothness.}
Large VLMs exhibit bounded Hessian spectra due to normalization layers,
residual connections, and pre-trained initialization.
Empirically we find the global Hessian spectral norm remains controlled.

\paragraph{Feature Regularity.}
CLIP encoders map to the unit sphere via normalization,
and are Lipschitz due to bounded Jacobians in ViT and ResNet blocks.

\subsection{Additional Remarks on the Robustness Decomposition}

The three terms in Theorem~\ref{thm:main} correspond to:

\begin{itemize}
\item \textbf{(A) Proximity to Prior:} Prevents drift from the learned manifold, 
safeguarding zero-shot transfer.
\item \textbf{(B) Parameter-Space Sharpness:} Controls curvature of the loss landscape 
and protects against adversarial perturbations.
\item \textbf{(C) Feature-Space Stability:} Ensures aligned class geometry across ID/OOD domains, 
enabling robustness to natural distribution shifts.
\end{itemize}

This decomposition motivates GRACE's design:
low-rank adaptive adversarial weight perturbation handles (B), 
proximity regularization handles (A), 
and cross-domain feature stabilization handles (C).

\section{Experimental Details for Empirical Validation of Failure Modes}
\label{app:geometry-details}

This appendix details the class-conditional metrics and curvature estimation used in Section~\ref{sec:geometry}.

\subsection{Feature-Space Geometry Metrics}
\label{app:feature-geometry-details}

\paragraph{Feature extraction.}
We use the penultimate image-encoder layer of CLIP ViT-B/32.
Each embedding $z(x)\in\mathbb{R}^d$ is $\ell_2$-normalized to lie on the unit hypersphere.
For each regime $r\!\in\!\{\text{ID},\text{OOD},\text{Adv}\}$ and class $c$, 
we collect 100 image embeddings and compute all statistics within-class.

\paragraph{Class-conditional cosine alignment.}
For each class $c$, we define domain-specific centroids
\[
\mu_r^c = \frac{1}{|D_r^c|}\sum_{x\in D_r^c} z(x).
\]
Alignment between domains is then given by
\[
\text{CS}_{\text{ID}\to\text{OOD}}^c = 
\langle \mu_{\text{ID}}^c, \mu_{\text{OOD}}^c\rangle,
\qquad
\text{CS}_{\text{ID}\to\text{Adv}}^c = 
\langle \mu_{\text{ID}}^c, \mu_{\text{Adv}}^c\rangle.
\]
We report their averages over classes.

\paragraph{Local Intrinsic Dimensionality (LID) and $\Delta$LID.}
LID measures local manifold complexity~\cite{ma2018characterizingadversarialsubspacesusing}.
For each feature $z_i$ within class $c$ and regime $r$, 
let $\{z_{i,j}\}_{j=1}^k$ be its $k{=}20$ nearest neighbors in cosine distance $r_{i,j}=1-\langle z_i,z_{i,j}\rangle$.
The LID estimate is
\[
\text{LID}(z_i)=
-\Big(\frac{1}{k}\sum_{j=1}^k\log\frac{r_{i,j}}{r_{i,k}}\Big)^{-1}.
\]
We compute the mean per-class, per-regime LID:
\[
\text{LID}_r^c = \frac{1}{|D_r^c|}\sum_{z_i\in D_r^c}\text{LID}(z_i).
\]
The class-conditional change in dimensionality relative to ID is:
\[
\Delta\text{LID}_{\text{ID}\to r}^c = \text{LID}_r^c - \text{LID}_{\text{ID}}^c,
\quad r\in\{\text{OOD},\text{Adv}\}.
\]
We report the class-averaged $\Delta\text{LID}$ values in Table~\ref{tab:lid}.
Lower $\Delta$LID indicates that the local manifold geometry is preserved under the shift.

\paragraph{PCA visualization.}
For qualitative plots (Figs.~\ref{fig:intro_analysis}, A2),
we fit a 3D PCA basis on pooled ID/OOD/Adv features from three random classes across methods and seeds,
and project all features into this common basis for visual comparison.

\subsection{Parameter-Space Curvature Estimation}
\label{app:curvature}

\paragraph{Hessian estimation.}
We estimate the top eigenvalue $\lambda_{\max}$ and normalized Frobenius norm $\|H\|_F/\sqrt{d}$ 
of the Hessian of the training loss $\mathcal{L}(\theta)$ 
with respect to all trainable parameters.
We use stochastic power iteration with $T{=}50$ iterations and batch size 512:
\[
v_{t+1} = \frac{H v_t}{\|H v_t\|_2},\quad
\lambda_{\max}\approx v_T^\top H v_T.
\]
We estimate $\|H\|_F$ via Hutchinson’s trace estimator with Gaussian probes $u_j\!\sim\!\mathcal{N}(0,I)$:
\[
\|H\|_F^2 \approx \frac{1}{m}\sum_{j=1}^m \|Hu_j\|_2^2.
\]
These metrics provide curvature proxies for model flatness and parameter-space complexity.

\paragraph{Loss-surface visualization.}
2D/3D loss slices (Fig.~\ref{fig:intro_analysis}(b), Fig.~A3) 
are obtained along two orthonormal perturbation directions $(\delta_1,\delta_2)$ 
in the flattened parameter vector, normalized to $\|\delta_i\|_2=1$,
visualizing $\mathcal{L}(\theta+\alpha\delta_1+\beta\delta_2)$ for $\alpha,\beta\in[-1,1]$.
Plots are qualitative and used only for illustration.

\begin{algorithm}[t]
\caption{\textbf{GRACE: Unified Robust Fine-Tuning with LoRA, LAR-AWP, and Gram Alignment}}
\label{alg:grace}
\begin{algorithmic}[1]
\Require Pretrained CLIP model $F_{\theta_0}$; training data $\mathcal{D}$; 
LoRA rank $r$; perturbation radius $\rho$; PGD budget $\epsilon$; 
tradeoff parameters $\lambda_{\text{LAR}}, \lambda_{\text{GV}}$.
\State Initialize LoRA parameters $\Theta = \{A_W, B_W\}_W$; freeze backbone weights.
\State Initialize layerwise AWP ranks $r_{\text{AWP}}^{(W)} \gets 0$ and curvature EMA $h_W \gets 0$ for all layers $W$.
\vspace{3pt}

\For{each training iteration}
    \State Sample minibatch $\{(x_i, y_i)\}_{i=1}^B$.
    \State Compute clean image features $f_{\text{ID}}(x_i)$ and task loss $\mathcal{L}_{\text{task}}$.
    \vspace{4pt}

    \Statex \textbf{// Step 1: Generate adversarial images}
    \State For each $i$, run PGD with radius $\epsilon$ to obtain
           $x_i^{\text{Adv}} \approx \arg\max_{\|\delta\| \le \epsilon}
           \mathcal{L}\big(F_{\theta}(x_i + \delta), y_i\big)$.
    \State Compute adversarial features $f_{\text{Adv}}(x_i)$ from $x_i^{\text{Adv}}$.
    \vspace{4pt}

    \Statex \textbf{// Step 2: LAR-AWP inner maximization (sharpness control)}
    \For{$t = 1$ to $T_{\text{AWP}}$}
        \State Update low-rank perturbation factors $(A_{\text{AWP}}, B_{\text{AWP}})$ by gradient ascent:
        \State \hspace{1em} $(A_{\text{AWP}}, B_{\text{AWP}})
            \gets (A_{\text{AWP}}, B_{\text{AWP}}) 
            + \eta \,\nabla_{\Delta} \mathcal{L}\big(F_{\theta,\Delta}(x_i^{\text{Adv}}), y_i\big)$.
        \State Project $\Delta$ onto the low-rank ball $\|\Delta\| \le \rho$ using layerwise rank masks $r_{\text{AWP}}^{(W)}$.
    \EndFor
    \State Compute perturbed features $f_{\text{AWP}}(x_i)$ using $W_{\text{pert}}(\theta,\Delta)$ in Eq.~\eqref{eq:lora-awp-param}.
    \vspace{4pt}

    \Statex \textbf{// Step 3: Gram-volume feature alignment}
    \State For each $i$, form Gram matrix $G_i$ from $\big(f_{\text{ID}}(x_i), f_{\text{Adv}}(x_i), f_{\text{AWP}}(x_i)\big)$ using Eq.~\eqref{eq:gram-matrix}.
    \State Compute Gram Alignment Loss $\mathcal{L}_{GV} = \sqrt{\lvert \det(G_i) \rvert}$.
    \vspace{4pt}

    \Statex \textbf{// Step 4: Curvature-based rank update (periodic)}
    \If{iteration $\bmod K = 0$}
        \State Estimate curvature proxy $h_W \approx \mathbb{E}\big[\|\nabla_W \mathcal{L}\|^2\big]$ using validation mini-batch gradients.
        \State Map $\{h_W\}_W$ to percentile bins and update layerwise ranks $r_{\text{AWP}}^{(W)}$:
        sharper layers $\Rightarrow$ higher $r_{\text{AWP}}^{(W)}$.
    \EndIf
    \vspace{4pt}

    \Statex \textbf{// Step 5: Outer minimization update}
    \State Update LoRA parameters $\Theta$ by minimizing
    \[
        \mathcal{L}_{\text{GRACE}} 
        = \mathcal{L}_{\text{task}}
        + \lambda_{\text{LAR}} \mathcal{L}_{\text{LAR-AWP}}
        + \lambda_{\text{GV}} \mathcal{L}_{\text{GV}}.
    \]
\EndFor
\end{algorithmic}
\end{algorithm}

\section{Method Details}
\label{app:method-details}

\subsection{LAR-AWP Rank Curriculum and Curvature Estimation}
\label{app:lar-awp-details}

Here we give the full description of the curvature estimator and rank curriculum
used in LAR-AWP (Section~\ref{subsec:lar-awp}). A central component of LAR-AWP is the assignment of layer-wise perturbation
ranks based on the local curvature of the loss landscape. Ideally, curvature
would be measured using the diagonal of the Hessian
$\mathrm{diag}(\nabla^2_{\!W}\mathcal{L}(\theta))$ for each weight matrix $W$.
However, computing or storing the exact Hessian is computationally infeasible
for CLIP-scale VLMs: a single iteration of finite-difference diagonal Hessian
estimation requires one backward pass per parameter, while estimating eigenvalues
by power iteration incurs prohibitive memory and compute overhead for hundreds
of millions of parameters.

\vspace{4pt}
\paragraph{Gauss--Newton Proxy.}
Following Sophia~\cite{sophia}, we adopt a tractable first-order proxy given by
the diagonal of the Gauss--Newton matrix. For a mini-batch
$\mathcal{B}_{\mathrm{val}}$ of size $n_v$, let
\[
g_W \;=\; \nabla_W 
\frac{1}{n_v} \sum_{(x_i,y_i)\in\mathcal{B}_{\mathrm{val}}}
\mathcal{L} \!\left(F_{\theta}(x_i),y_i\right).
\]
The curvature estimator is then defined as
\begin{equation}
\widehat{h}_W 
\;=\; 
n_v \, (g_W \odot g_W),
\label{eq:app-curvature-proxy}
\end{equation}
where $\odot$ denotes elementwise multiplication.
For common losses such as cross-entropy, the expectation of $(g_W\odot g_W)$
approximates the diagonal of the Gauss--Newton matrix 
$H_{\mathrm{GN}} \preceq \nabla^2\mathcal{L}$.
Therefore, $\widehat{h}_W$ is a \emph{biased underestimator} of the true Hessian
diagonal: it systematically underestimates curvature, but preserves the relative
ordering of layers.

\vspace{4pt}
\paragraph{Bias and Its Impact on Rank Adaptation.}
LAR-AWP does not require an unbiased estimate of curvature.
Rather, it requires a stable \emph{relative} ordering of layers by sharpness.
Since the Gauss--Newton proxy shrinks curvature uniformly but monotonically,
layers with higher true curvature still satisfy
$\widehat{h}_{W_1} > \widehat{h}_{W_2}$ whenever
$H_{W_1} \succ H_{W_2}$.
Rank assignment is based on curvature percentiles:
\[
r_{\mathrm{AWP}}
\;\propto\;
\mathbf{1}\left\{\widehat{h}_W \ge \tau_{p}\right\}, 
\qquad
\tau_{p} = \mathrm{quantile}\!\left(\{\widehat{h}_W\}_W,\,p\right),
\]
with $p=0.8$ in our implementation.
Thus, any consistent monotonic bias is harmless: only the ranking matters.

\vspace{4pt}
\paragraph{Stabilization via Exponential Moving Average.}
To mitigate mini-batch noise, we maintain an EMA of the curvature proxy:
\begin{equation}
h_W^{(t)}
\;=\;
\beta \, h_W^{(t-1)}
+
(1-\beta)\, \widehat{h}_W^{(t)},
\qquad
\beta \in [0.85, 0.95].
\label{eq:app-curvature-ema}
\end{equation}
We also normalize by parameter size to avoid scale inflation in large matrices:
\[
h_W^{\text{norm}} 
= \frac{1}{|W|} \sum \widehat{h}_W.
\]
Curvature updates are performed once every $K$ iterations ($K=1000$ by default),
balancing responsiveness and stability.

\vspace{4pt}
\paragraph{Practical Rank Allocation.}
Given the stabilized curvature values $\{h_W\}$, we assign AWP ranks via a
piecewise mapping:
sharp layers (top 20\%) receive the highest rank,
moderately curved layers receive intermediate ranks,
and flat layers (bottom 20\%) receive minimal or zero rank.
This curriculum focuses smoothing where the loss landscape is most anisotropic.

\subsection{Using AWP as a Proxy for Out-of-Distribution Robustness}
\label{app:awp-ood}

A key motivation behind LAR-AWP is the observation that
\emph{weight-space adversarial perturbations provide a tractable proxy for
natural distribution shifts}. In this subsection, we formalize this connection
and provide empirical support for the use of AWP as an OOD surrogate in VLM
fine-tuning.

\vspace{4pt}
\paragraph{Background: OOD shifts as structured adversarial perturbations.}
Recent studies have suggested that natural distribution shifts often correspond
to \emph{structured, low-dimensional perturbations} in the feature space of
pre-trained models. In CLIP-like VLMs, these shifts manifest primarily as
changes in texture, lighting, occlusion statistics, or object context.
Such variations induce predictable deformations in the intermediate
representations of the visual encoder that can be approximated by
\emph{parametric perturbations of the model weights}.

Formally, let $P_{\mathrm{ID}}$ and $P_{\mathrm{OOD}}$ denote the ID and OOD
input distributions. For a model $f_{\theta}$, the OOD-induced representation
drift for sample $x$ is
\[
\Delta_{\mathrm{OOD}}(x)
=
f_{\theta}(x) - f_{\theta}(x'),
\qquad
x' \sim P_{\mathrm{OOD}}(\cdot \mid x),
\]
where $x'$ denotes an OOD variant of $x$ (e.g., ImageNet-R, -V2, -S, A/A+).
Empirically, $\Delta_{\mathrm{OOD}}(x)$ lies in a low-dimensional subspace
associated with sharp, anisotropic directions of the weight-loss landscape.
This provides the bridge to weight-space adversarial perturbations.

\vspace{4pt}
\paragraph{AWP as a local surrogate for OOD feature drift.}
Adversarial Weight Perturbation (AWP)~\cite{awp} searches for a weight
perturbation $\Delta$ that maximally increases loss on a given input:
\[
\Delta_{\mathrm{AWP}}
=
\arg\max_{\|\Delta\|\le\rho}
\mathcal{L}\big(f_{\theta+\Delta}(x),y\big).
\]
In high-dimensional VLMs, the maximizer $\Delta_{\mathrm{AWP}}$ tends to align
with directions of high curvature, i.e., 
eigenvectors corresponding to large eigenvalues of $\nabla^2_{\!\theta}\mathcal{L}$.
These directions correspond to
feature-space instabilities that are \emph{also} amplified under natural OOD
shifts. Thus,
\[
\Delta_{\mathrm{AWP}}(x)
\;\parallel\; \Delta_{\mathrm{OOD}}(x)
\qquad
\text{(up to a monotone scaling factor)}.
\]
This alignment implies that AWP perturbs the model along directions that mimic
OOD-induced representation drift.

\vspace{4pt}
\paragraph{Why AWP captures OOD geometry in CLIP.}
Pre-trained CLIP encoders exhibit highly anisotropic curvature:
a small fraction of layers (often the late transformer blocks)
dominate the Hessian spectrum.
These layers are also empirically the most sensitive to OOD corruption,
as shown in our layerwise curvature analysis
(App.~\ref{app:layerwise-curvature}).
Therefore, weight-space adversarial perturbations naturally
concentrate in the same regions of parameter space that dominate OOD behavior.

In GRACE, the rank-adaptive LAR-AWP module explicitly targets these layers,
ensuring that adversarial perturbations span the high-curvature subspace where
OOD feature drift resides.

\vspace{4pt}
\paragraph{Empirical evidence: AWP feature drift correlates with OOD drift.}
We measure the correlation between:
(i) feature displacement induced by AWP, and  
(ii) feature displacement induced by real OOD samples.

Let

\begin{figure}
    \centering
    \includegraphics[width=\linewidth]{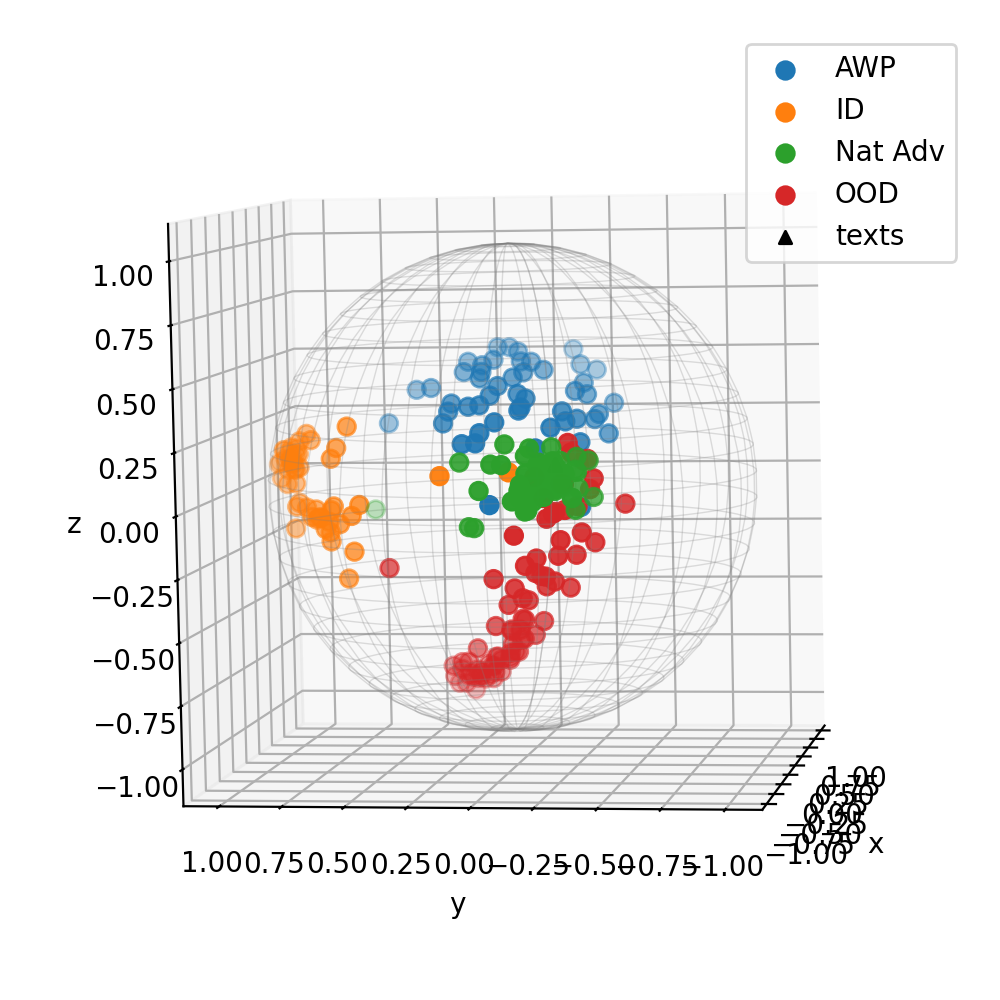}
    \caption{\textbf{3D visualization of CLIP feature geometry under different shifts.} Using normalized embeddings projected onto the unit sphere, we compare ID samples, natural adversarial (Nat Adv) variants, OOD samples, and features obtained under AWP perturbations. AWP produces feature displacements that closely follow the structure of real OOD and natural adversarial shifts, confirming that curvature-aligned weight perturbations serve as an effective proxy for robustness-critical distribution drift.}

    \label{fig:awp_ood}
\end{figure}

\begin{dmath}
d_{\mathrm{AWP}}(x)
    = \| f_{\theta}(x) - f_{\theta+\Delta_{\mathrm{AWP}}}(x) \|_2,
\qquad
d_{\mathrm{OOD}}(x)
    = \mathbb{E}_{x'\sim P_{\mathrm{OOD}}}[ \| f_{\theta}(x) - f_{\theta}(x') \|_2 ].
\end{dmath}
Across ImageNet-V2/S/R and ImageNet-A/A+, we observe a strong positive
correlation (Table \ref{tab:awp_ood}, Figure ~\ref{fig:awp_ood}), confirming that AWP and OOD shifts
induce similar instability patterns in CLIP's feature space.
Layers with large curvature exhibit the strongest coupling.

\begin{table}[t]
    \centering
    \small

    \label{tab:geometry_metrics}
        \centering
        \begin{tabular}{lcc}
            \toprule
            Shift & Cosine Similarity & L2 Distance \\
            \midrule
            ID$\to$OOD & 0.35 & 6.61 \\
            ID$\to$AWP & 0.32 & 6.87 \\
            \bottomrule
        \end{tabular}
        \caption{\textbf{Feature-space displacement induced by OOD data and AWP.}  We report cosine similarity and L2 feature distance between ID representations and those produced under OOD samples or adversarial weight perturbations. The similar displacement magnitudes show that AWP produces feature shifts comparable to true OOD drift, supporting its utility as a proxy for robustness stress-testing.}

        \label{tab:awp_ood}
\end{table}

\vspace{4pt}
\paragraph{Implication for training.}
Because AWP tracks OOD-sensitive directions, optimizing robustness against AWP
implicitly improves generalization on OOD benchmarks:
\[
\min_\theta \max_{\Delta}
\mathcal{L}(f_{\theta+\Delta})
\quad\Longrightarrow\quad
\min_\theta
\max_{x'\sim P_{\mathrm{OOD}}}
\mathcal{L}(f_{\theta}(x')).
\]
This explains why GRACE improves both adversarial and OOD robustness, even
though OOD samples are \emph{not} used during training.

\paragraph{Conclusion.}
LAR-AWP acts as a computationally efficient surrogate for natural distribution
shifts in CLIP. By perturbing the model along high-curvature directions that
correlate with OOD drift, GRACE enforces stability in precisely those regions
of parameter space where OOD generalization typically fails.

\subsection{Training Algorithm}
\label{app:training-details}

Algorithm \ref{alg:grace} presents full pseudocode for GRACE, including adversarial example generation,
inner LAR-AWP steps, Gram-volume alignment loss computation,
and outer optimization of the LoRA parameters $\Theta$.

\section{Extended Experimental Details}
\label{app:exp_details}

\subsection{Preprocessing and Prompts}
We use CLIP preprocessing for ViT-B/32. For zero-shot evaluations, we use the standard ImageNet prompt set and a single fixed template set for transfer datasets (no ensembling unless stated).

\subsection{Datasets}
In this section, we provide additional details for the datasets used for the fine-tuning experiments in the paper. We employ the ImageNet \cite{imagenet} dataset along with its variants for the fine-tuning and OOD experiments in the paper. 

\begin{figure}[h]
    \centering
    \includegraphics[width=\linewidth]{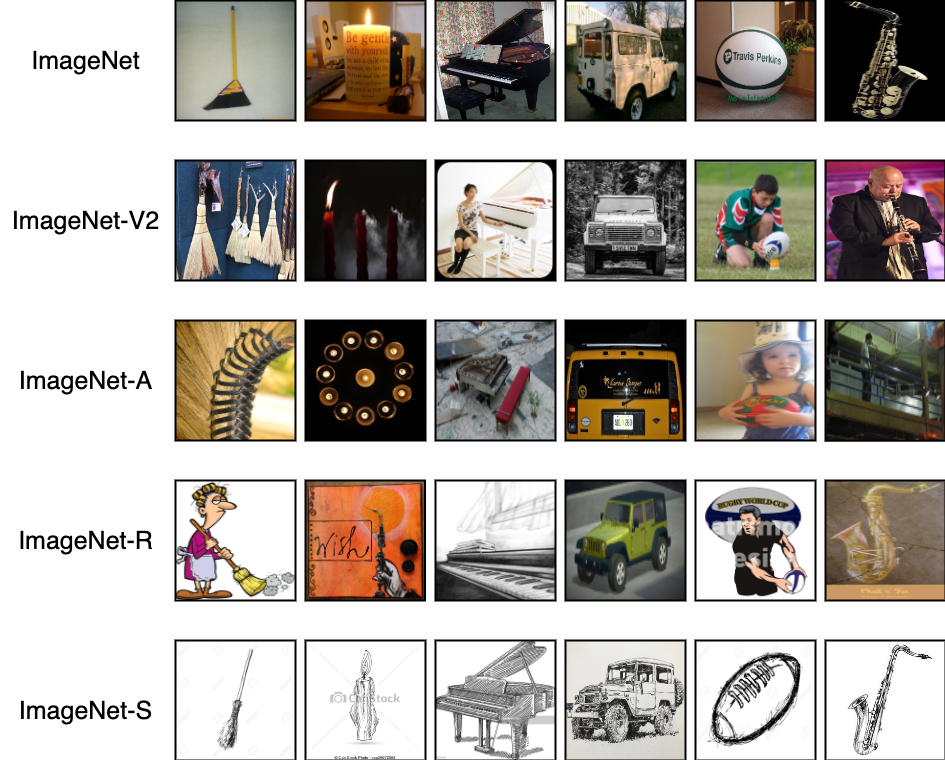}
    \caption{Samples from the ImageNet dataset}
    \label{fig:imagenet}
\end{figure}

The ImageNet \cite{imagenet} dataset consists of images categorized into 1000 classes. To test the OOD robustness of the fine-tuned models, we use the following related datasets:  ImageNet-V2 \cite{imagenetv2}, ImageNet-Rendition \cite{imagenet-r}, ImageNet-Adversarial \cite{imagenet-a}, and ImageNet-Sketch \cite{imagenet-s}. These datasets are considered as natural distribution shifts of the ImageNet dataset, hence we use these datasets to evaluate the robustness of models fine-tuned on ImageNet using these datasets. Figure \ref{fig:imagenet} shows samples from all these datasets.

\subsection{Training Details}
The code for the main experiments is built on PyTorch \cite{pytorch}, and Transformers \cite{transformers_lib} libraries. All the experiments in the paper were done on a SLURM cluster with 8 NVIDIA A40 GPUs per node.

\subsection{Pre-Trained Weights}
Throughout the experiments, we employed the pre-trained weights provided by OpenAI. These weights can be accessed through the following URLs:

\begin{itemize}
    \item ViT-B/32: \href{https://huggingface.co/openai/clip-vit-base-patch32}{https://huggingface.co/openai/clip-vit-base-patch32}
    \item ViT-B/16: \href{https://huggingface.co/openai/clip-vit-base-patch16}{https://huggingface.co/openai/clip-vit-base-patch16}
    \item ViT-L/14: \href{https://huggingface.co/openai/clip-vit-large-patch14}{https://huggingface.co/openai/clip-vit-large-patch14}
\end{itemize}

\subsection{Hyperparameters}
All the hyper-parameters used in the fine-tuning experiments were selected based on the ID validation sets for each dataset.  We use validation set provided. The corresponding values selected for the hyper-parameters are shown in Table \ref{tab:hyperparam}.

\begin{table}[h]
    \centering
    \resizebox{\linewidth}{!}{
    \begin{tabular}{c|c}
        \hline
        \textbf{Hyperparameter} & \textbf{Selected Value} \\
        \hline
        Rank of Weight Updates ($r_W$) & 64 \\
        Maximum Rank of Weight Perturbations ($r_{AWP})$ & 4 \\
        Input Perturbation Strength ($\epsilon$) & $4/255$ \\
        Input Perturbation Step Size ($\eta_1$) & $1/255$ \\
        Learning Rate ($\eta_3$) & 2e-4 \\
        Gradient Threshold Percentile for Weight Perturbations ($\varphi_{AWP}$) & 80 \\ 
        Number of Epochs & 50 \\
        Batch Size & 256 \\
        \hline
    \end{tabular}
    }
    \caption{List of hyperparameter values used for the fine-tuning experiments.}
    \label{tab:hyperparam}
\end{table}

\section{Additional Discussion}

\subsection{Analyzing AWP Ranks}
\label{app:awp-rank-annealing}

\begin{figure}[h]
    \centering
    \includegraphics[width=\linewidth]{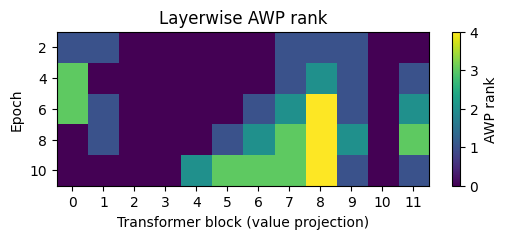}
    \caption{\textbf{Evolution of Layerwise AWP Rank During Training.} The heatmap shows the adaptive adversarial weight perturbation (AWP) rank assigned to each transformer block across training epochs. }
    \label{fig:awp-rank-progression}
\end{figure}

To understand how GRACE allocates adversarial perturbation capacity across the network, we analyze the temporal progression of the learned AWP rank $r^{(\ell)}_{\text{AWP}}$ and relate it to the curvature structure of the CLIP ViT-B/32 encoder. Each layer begins with rank~0, and the rank is subsequently increased or decreased throughout training based on the layer’s instantaneous curvature estimate. This allows us to directly visualize how GRACE responds to the evolving geometry of the model.

\paragraph{Curvature anisotropy shapes where ranks appear.}
Figure~\ref{fig:layerwise-curvature} shows that curvature is highly anisotropic: mid-to-deep layers (blocks~4--8) exhibit a pronounced curvature peak, while early and late layers remain substantially flatter. This curvature pattern precisely mirrors the region where AWP ranks emerge. As seen in Fig.~\ref{fig:awp-rank-progression}, only layers within this high-curvature band ever receive non-zero ranks; layers with consistently low curvature (blocks~0--3 and block~11) retain a rank of zero throughout all epochs.

\vspace{4pt}
\paragraph{Observation: Rank progression tracks curvature dynamics.}
The temporal evolution of AWP ranks exhibits a clear structural correspondence with the curvature plot:
\begin{enumerate}[leftmargin=*]
    \item \textbf{High-curvature layers accumulate rank.}  
    In blocks~4--9, where curvature is large in Fig.~\ref{fig:layerwise-curvature}, ranks progressively rise from~0 to values between~1 and~4.
    \item \textbf{Rank magnitude aligns with curvature magnitude.}  
    The maximum rank (4) appears in block~8—the same layer with the highest curvature.
    \item \textbf{Fluctuations reflect curvature changes over training.}  
    Layers whose curvature decreases over epochs exhibit corresponding decreases in AWP rank (e.g., blocks~6--7), whereas layers with persistent curvature peaks (e.g., block~8) maintain higher ranks for longer.
\end{enumerate}
Thus, the progression in Fig.~\ref{fig:awp-rank-progression} is not arbitrary: it faithfully traces the curvature landscape and its evolution over training.

\vspace{4pt}
\paragraph{Implications.}
This analysis confirms that GRACE allocates adversarial weight perturbations exactly where the model is geometrically sensitive. The AWP rank increases when a layer becomes sharp and decreases when the layer is flattened by GRACE’s optimization dynamics. Conversely, layers that remain flat never accumulate rank, ensuring that perturbation capacity is used selectively and efficiently.

\vspace{4pt}
\paragraph{Summary.}
The AWP rank heatmap closely follows the curvature profile of the encoder. High-curvature layers receive increasing rank, low-curvature layers retain zero rank, and fluctuations in curvature over training are mirrored by fluctuations in AWP rank. This provides direct empirical evidence that GRACE’s layerwise perturbation allocation is governed by the geometry of the model.

\subsection{Hyperparameter Analysis}
In order to understand the impact of various hyperparameter choices on the downstream task performance of our approach, we performed experiments varying the various hyperparameters in our approach. In our experiments, we varied the following hyperparameters: (1) rank of weight updates $\{ 16,32,64 \}$ (2) maximum rank of weight perturbation $\{ 2, 4, 8 \}$ (3) input perturbation strength $\{ 1/255, 2/255, 4/255 \}$ (4) input perturbation step size $\{ 1/255, 2/255, 4/255 \}$ (5) gradient threshold percentile $\{ 40, 60, 80 \}$. The final results for the hyperparameter analysis have been summarized in Table \ref{tab:hyperparameter_analysis}. All the results were for a ViT-B/32 CLIP model finetuned on the ImageNet Dataset and tested on other domains.

\begin{table*}[t]
    \centering
    \tabcolsep=2.7pt
    \extrarowheight=2pt
    \resizebox{0.8\linewidth}{!}{
    \begin{tabular}{c|c|c|c|c|c|c|c|c|c|c}
        \toprule
       \multirow{3}{*}{\textbf{Hyperparameter}} & 
       \multirow{3}{*}{\textbf{Value}} & 
       \multirow{3}{*}{\textbf{ID}} & 
       \multicolumn{2}{c|}{\textbf{Out-Of-Distribution}} &
       \multicolumn{4}{c|}{\textbf{Adversarial}} &
       \multicolumn{2}{c}{\textbf{Statistics}} \\
        \cmidrule(l{0.6em}r{0.6em}){4-5}
        \cmidrule(l{0.6em}r{0.6em}){6-9}
        \cmidrule(l{0.6em}r{0.6em}){10-11}
         & & & Domain Shift & Zero Shot & Adv ID & Adv OOD  & Adv ZS  & Natural Adv & Avg OOD & Avg Adv \\
         \toprule
         \midrule
         
         \multirow{3}{*}{Rank of Weight Updates $(r_W)$}
           & 16 & 72.80 & 53.10 & 58.10 & 22.10 & 18.30 & 18.90 & 21.10 & 55.50 & 20.10 \\
           & 32 & 73.60 & 53.80 & 59.00 & 23.90 & 19.40 & 20.30 & 22.00 & 56.50 & 21.20 \\
           & 64 & 74.21 & 54.41 & 59.61 & 25.44 & 20.55 & 21.10 & 22.67 & 57.01 & 22.44 \\
         \cmidrule{1-11}\noalign{\vskip 0.5ex}

         \multirow{3}{*}{Maximum Rank of Weight Perturbations $(r_{\text{AWP}})$}
           & 2 & 74.30 & 54.90 & 59.70 & 23.10 & 18.90 & 19.40 & 21.90 & 56.50 & 20.80 \\
           & 4 & 74.21 & 54.41 & 59.61 & 25.44 & 20.55 & 21.10 & 22.67 & 57.01 & 22.44 \\
           & 8 & 73.70 & 53.90 & 58.90 & 26.80 & 21.20 & 22.10 & 23.30 & 56.30 & 23.10 \\
         \cmidrule{1-11}\noalign{\vskip 0.5ex}

         \multirow{3}{*}{Input Perturbation Strength $(\epsilon)$}
           & 1/255 & 75.10 & 55.20 & 60.10 & 18.70 & 14.40 & 15.80 & 21.40 & 57.00 & 17.60 \\
           & 2/255 & 74.70 & 54.80 & 59.80 & 22.30 & 18.00 & 18.90 & 22.10 & 56.90 & 19.70 \\
           & 4/255 & 74.21 & 54.41 & 59.61 & 25.44 & 20.55 & 21.10 & 22.67 & 57.01 & 22.44 \\
         \cmidrule{1-11}\noalign{\vskip 0.5ex}

         \multirow{3}{*}{Input Perturbation Step Size $(\eta)$}
           & 1/255 & 74.21 & 54.41 & 59.61 & 25.44 & 20.55 & 21.10 & 22.67 & 57.01 & 22.44 \\
           & 2/255 & 73.90 & 54.10 & 59.20 & 24.90 & 20.10 & 20.70 & 22.40 & 56.70 & 21.90 \\
           & 4/255 & 73.40 & 53.50 & 58.80 & 23.20 & 19.00 & 19.10 & 21.90 & 56.10 & 20.40 \\
         \cmidrule{1-11}\noalign{\vskip 0.5ex}

         \multirow{3}{*}{Gradient Threshold Percentile $(\varphi_{\text{AWP}})$}
           & 40 & 75.00 & 54.90 & 60.10 & 20.90 & 16.50 & 17.40 & 21.70 & 56.70 & 19.10 \\
           & 60 & 74.60 & 54.60 & 59.80 & 23.40 & 18.90 & 20.10 & 22.30 & 56.90 & 20.80 \\
           & 80 & 74.21 & 54.41 & 59.61 & 25.44 & 20.55 & 21.10 & 22.67 & 57.01 & 22.44 \\

        \bottomrule
    \end{tabular}
    }
    \caption{\textbf{Hyperparameter sensitivity results for GRACE (ViT-B/32).} CLIP ViT-B/32 finetuned on ImageNet \cite{imagenet} dataset and evaluated on ImageNet(variants). The numbers represent top-1 accuracy. }
    \label{tab:hyperparameter_analysis}
\end{table*}

\begin{table*}[t]
    \centering
    \tabcolsep=2.7pt
    \extrarowheight=2pt
    \resizebox{0.8\linewidth}{!}{
    \begin{tabular}{c|c|c|c|c|c|c|c|c|c}
        \toprule
       \multirow{3}{*}{\textbf{Method}} & \multirow{3}{*}{\textbf{ID}} & \multicolumn{2} {c|} {\textbf{Out-Of-Distribution}} & \multicolumn{4} {c|} {\textbf{Adversarial}} & \multicolumn{2} {c} {\textbf{Statistics}} \\
        \cmidrule(l{0.6em}r{0.6em}){3-4} \cmidrule(l{0.6em}r{0.6em}){5-8} \cmidrule(l{0.6em}r{0.6em}){9-10}
         &  & Domain Shift & Zero Shot & Adv ID & Adv OOD  & Adv ZS  & Natural Adv & Avg OOD & Avg Adv \\
         &  & (Same Classes) & (Unseen Classes) & ($\epsilon{=}4/255$) & ($\epsilon{=}4/255$) & ($\epsilon{=}4/255$) & ImageNet-A/A-Plus &  & \\
         \toprule
        \midrule
          CLIP & 68.35 & 62.60 & 64.92 & 0.00 & 0.00 & 0.13 & 52.48 & 63.76 & 13.15 \\
          \cmidrule{1-10}\noalign{\vskip 0.5ex}
          Vanilla FT & 81.30 & 60.43 & 71.48 & 0.00 & 0.00 & 0.00 & 39.35 & 65.96 & 9.84 \\
          WISE-FT \cite{wiseft} & 82.50 & 66.60 & 68.68 & 0.00 & 0.00 & 0.00 & 49.95 & 67.64 & 12.49 \\
          FLYP \cite{flyp} & 82.60 & 64.67 & 69.91 & 0.00 & 0.00 & 0.00 & 50.85 & 67.29 & 12.71 \\
          TPGM \cite{tpgm} & 82.85 & 65.50 & 70.55 & 0.00 & 0.00 & 0.00 & 50.85 & 68.03 & 12.71 \\
          SPD \cite{spd} & 82.40 & 64.90 & 70.06 & 0.00 & 0.00 & 0.00 & 50.35 & 67.48 & 12.59 \\
          \cmidrule{1-10}\noalign{\vskip 0.5ex}
          TeCoA \cite{tecoa} & 66.80 & 47.20 & 49.15 & 31.20 & 21.90 & 35.13 & 11.65 & 48.18 & 24.97 \\
          FARE \cite{fare} & 58.20 & 48.17 & 59.66 & 18.50 & 16.20 & 23.04 & 12.40 & 53.92 & 17.54 \\
          PMG-AFT \cite{pmg_aft} & 68.50 & 51.47 & 53.37 & 33.60 & 23.87 & 38.51 & 15.40 & 52.42 & 27.85 \\
          LAAT \cite{laat} & 64.20 & 49.07 & 51.55 & 27.80 & 20.43 & 36.65 & 20.10 & 50.31 & 26.25 \\
          \midrule
          \rowcolor{gray!15}
          GRACE (Ours) & 82.30 & 63.10 & 71.48 & 32.40 & 25.97 & 41.39 & 34.00 & 67.29 & 33.44 \\
          \bottomrule
    \end{tabular}
    }
    \caption{\textbf{Unified summary across settings (ViT-B/16).} \emph{ID} uses ImageNet clean; \emph{Domain Shift (Same Classes)} is OOD Avg over ImageNet-V2/S/R (clean); \emph{Zero Shot} is ZS Avg (clean) across 8 datasets; \emph{Targeted Adversarial} reports (ID/OOD/ZS) accuracies under AutoAttack (APGD-CE, $\epsilon{=}4/255$); \emph{Natural Adversarial} is the clean average over ImageNet-A/A-Plus.\vspace{-10pt}}
    \label{tab:summary_b_16}
\end{table*}

\begin{table*}[]
    \centering
    \tabcolsep=3pt
    \extrarowheight=2pt
    \resizebox{0.8\linewidth}{!}{
    \begin{tabular}{c | c  c | c  c | c  c | c  c | c  c | c  c | c  c | c  c }
    \toprule
         & \multicolumn{2} {c|} {\textbf{ID}} & \multicolumn{6} {c|} {\textbf{Domain Shift}} & \multicolumn{4} {c|} {\textbf{Natural Adversarial}} & \multicolumn{4} {c} {\textbf{Statistics}}\\
         \cmidrule(l{0.6em}r{0.6em}){4-9} \cmidrule(l{0.6em}r{0.6em}){10-13} \cmidrule(l{0.6em}r{0.6em}){14-17}
         \textbf{Method}  & \multicolumn{2} {c|} {\textbf{ImageNet}} & \multicolumn{2} {c|} {\textbf{ImageNet-V2}} & \multicolumn{2} {c|} {\textbf{ImageNet-S}} & \multicolumn{2} {c|} {\textbf{ImageNet-R}} & \multicolumn{2} {c|} {\textbf{ImageNet-A}} & \multicolumn{2} {c|} {\textbf{ImageNet-A-Plus}} & \multicolumn{2} {c|} {\textbf{OOD Avg.}}  & \multicolumn{2} {c} {\textbf{Nat Adv Avg}}\\

         & Clean & Adv & Clean & Adv & Clean & Adv & Clean & Adv & Clean & Adv & Clean & Adv & Clean & Adv & Clean & Adv \\
    \toprule
    \midrule
     CLIP \cite{clip} & 68.35 & 0.00 & 61.90 & 0.00 & 48.30 & 0.00 & 77.60 & 0.01 & 50.10 & 0.00 & 54.85 & 0.00 & 62.60 & 0.00 & 52.48 & 0.00 \\
     \cmidrule{1-17}\noalign{\vskip 0.5ex}
     Vanilla FT & 81.30 & 0.00 & 70.60 & 0.00 & 45.10 & 0.00 & 65.60 & 0.00 & 36.60 & 0.00 & 42.10 & 0.00 & 60.43 & 0.00 & 39.35 & 0.00 \\
     WISE-FT \cite{wiseft} & 82.50 & 0.00 & 73.10 & 0.00 & 51.60 & 0.00 & 75.10 & 0.00 & 47.60 & 0.00 & 52.30 & 0.00 & 66.60 & 0.00 & 49.95 & 0.00 \\
      FLYP \cite{flyp} & 82.60 & 0.00 & 73.00 & 0.00 & 49.60 & 0.00 & 71.40 & 0.01 & 48.10 & 0.00 & 53.60 & 0.00 & 64.67 & 0.00 & 50.85 & 0.00 \\
      TPGM \cite{tpgm} & 82.85 & 0.00 & 73.50 & 0.00 & 50.20 & 0.00 & 72.80 & 0.00 & 48.50 & 0.00 & 53.20 & 0.00 & 65.50 & 0.00 & 50.85 & 0.00 \\
      SPD \cite{spd} & 82.40 & 0.00 & 72.80 & 0.00 & 49.80 & 0.00 & 72.10 & 0.00 & 47.90 & 0.00 & 52.80 & 0.00 & 64.90 & 0.00 & 50.35 & 0.00 \\
      \cmidrule{1-17}\noalign{\vskip 0.5ex}
      TeCoA \cite{tecoa} & 66.80 & 31.20 & 52.30 & 22.50 & 32.10 & 16.80 & 57.20 & 26.40 & 8.50 & 1.20 & 14.80 & 2.10 & 47.20 & 21.90 & 11.65 & 1.65 \\
      FARE \cite{fare} & 58.20 & 18.50 & 49.80 & 12.60 & 35.40 & 14.20 & 59.30 & 21.80 & 9.20 & 0.90 & 15.60 & 1.40 & 48.17 & 16.20 & 12.40 & 1.15 \\
      PMG-AFT \cite{pmg_aft} & 68.50 & 33.60 & 54.80 & 25.20 & 38.40 & 18.50 & 61.20 & 27.90 & 12.30 & 1.85 & 18.50 & 3.20 & 51.47 & 23.87 & 15.40 & 2.53 \\
      LAAT \cite{laat} & 64.20 & 27.80 & 51.50 & 20.30 & 36.80 & 15.60 & 58.90 & 25.40 & 17.80 & 3.10 & 22.40 & 4.80 & 49.07 & 20.43 & 20.10 & 3.95 \\
      \midrule
      \rowcolor{gray!15}
      GRACE & 82.30 & 32.40 & 72.20 & 26.10 & 49.30 & 22.30 & 67.80 & 29.50 & 31.20 & 5.20 & 36.80 & 7.30 & 63.10 & 25.97 & 34.00 & 6.25 \\

    \bottomrule

    \end{tabular}
    }
    \caption{\textbf{OOD Results on ImageNet (ViT-B/16).} CLIP ViT-B/16 finetuned on ImageNet \cite{imagenet} dataset and evaluated on ImageNet variants. The numbers are top-1 accuracy (\%). OOD Avg averages ImageNet-V2, -S, -R; Nat Adv Avg averages ImageNet-A and A-Plus.}
    \label{tab:imagenet_vit_b_16}
\end{table*}

\begin{table*}[]
    \centering
    \resizebox{\linewidth}{!}{
    \begin{tabular}{c|cc|cc|cc|cc|cc|cc|cc|cc|cc}
    \toprule
         & \multicolumn{16} {c|} {\textbf{Zero shot Datasets}} & \multicolumn{2} {c} {\textbf{Statistics}}\\
         \cmidrule(l{0.6em}r{0.6em}){2-17} \cmidrule(l{0.6em}r{0.6em}){18-19}
        & \multicolumn{2} {c|} {\textbf{CalTech101}} & \multicolumn{2} {c|} {\textbf{Cars}} & \multicolumn{2} {c|} {\textbf{DTD}} & \multicolumn{2} {c|} {\textbf{EuroSAT}} & \multicolumn{2} {c|} {\textbf{FGVC}} & \multicolumn{2} {c|} {\textbf{Flowers}} & \multicolumn{2} {c|} {\textbf{Oxford Pets}} & \multicolumn{2} {c|} {\textbf{STL-10}} & \multicolumn{2} {c} {\textbf{ZS Avg.}} \\
        Method & Clean & Adv & Clean & Adv & Clean & Adv & Clean & Adv & Clean & Adv & Clean & Adv & Clean & Adv & Clean & Adv & Clean & Adv \\
    \toprule
    \midrule
     CLIP  & 91.48 & 0.41 & 59.74 & 0.00 & 43.85 & 0.00 & 54.70 & 0.00 & 18.96 & 0.00 & 67.42 & 0.00 & 86.13 & 0.00 & 97.10 & 0.64 & 64.92 & 0.13 \\
     \cmidrule{1-19}\noalign{\vskip 0.5ex}
      Vanilla FT & 93.20 & 0.00 & 68.40 & 0.00 & 52.30 & 0.00 & 68.50 & 0.00 & 26.80 & 0.00 & 75.60 & 0.00 & 90.20 & 0.00 & 96.80 & 0.00 & 71.48 & 0.00 \\
      WISE-FT & 92.80 & 0.00 & 65.30 & 0.00 & 48.60 & 0.00 & 62.40 & 0.00 & 22.50 & 0.00 & 71.80 & 0.00 & 88.60 & 0.00 & 97.40 & 0.00 & 68.68 & 0.00 \\
      FLYP  & 93.10 & 0.00 & 66.80 & 0.00 & 50.20 & 0.00 & 64.80 & 0.00 & 24.30 & 0.00 & 73.40 & 0.00 & 89.50 & 0.00 & 97.20 & 0.00 & 69.91 & 0.00 \\
      TPGM  & 93.50 & 0.00 & 67.20 & 0.00 & 51.40 & 0.00 & 66.10 & 0.00 & 25.10 & 0.00 & 74.20 & 0.00 & 89.80 & 0.00 & 97.10 & 0.00 & 70.55 & 0.00 \\
      SPD  & 93.00 & 0.00 & 66.50 & 0.00 & 50.80 & 0.00 & 65.30 & 0.00 & 24.70 & 0.00 & 73.80 & 0.00 & 89.40 & 0.00 & 97.00 & 0.00 & 70.06 & 0.00 \\
      \cmidrule{1-19}\noalign{\vskip 0.5ex}
      TeCoA  & 84.63 & 69.07 & 37.50 & 24.03 & 28.93 & 18.16 & 38.60 & 23.48 & 7.81 & 3.13 & 36.13 & 23.05 & 70.31 & 45.70 & 89.25 & 74.41 & 49.15 & 35.13 \\
      FARE  & 88.20 & 42.50 & 52.40 & 15.20 & 38.50 & 12.80 & 48.60 & 14.50 & 15.30 & 2.10 & 58.40 & 16.30 & 82.10 & 28.60 & 93.80 & 52.30 & 59.66 & 23.04 \\
      PMG-AFT  & 86.50 & 71.20 & 42.30 & 28.50 & 34.38 & 22.15 & 44.20 & 27.80 & 10.25 & 4.85 & 42.96 & 26.40 & 75.00 & 50.20 & 91.40 & 76.85 & 53.37 & 38.51 \\
      LAAT  & 85.80 & 68.40 & 40.20 & 26.30 & 32.10 & 20.50 & 41.30 & 25.60 & 9.20 & 4.10 & 40.10 & 24.80 & 73.20 & 48.30 & 90.50 & 75.20 & 51.55 & 36.65 \\
      \midrule
      \rowcolor{gray!15}
      GRACE & 93.60 & 72.80 & 68.90 & 32.40 & 52.60 & 25.30 & 67.20 & 31.50 & 26.40 & 6.20 & 75.80 & 29.80 & 90.40 & 54.60 & 96.90 & 78.50 & 71.48 & 41.39 \\
    \bottomrule
    \end{tabular}
    }
    \caption{\textbf{Clean and adversarial evaluation on zero-shot image classification datasets (ViT-B/16).} Models are trained on ImageNet; all other datasets are zero-shot. ZS Avg is the mean across the 8 datasets.}
    \label{tab:zero_shot_vit_b_16}
\end{table*}